\documentclass[10pt,twocolumn,letterpaper]{article}

\usepackage{iccv}
\usepackage{times}
\usepackage{epsfig}
\usepackage{graphicx}
\usepackage{amsmath}
\usepackage{amssymb}
\usepackage{subfig}
\usepackage{mathrsfs}
\usepackage{blindtext}
\usepackage[english]{babel}
\usepackage{color}
\usepackage[noend]{algpseudocode}
\usepackage{algorithmicx,algorithm}
\usepackage[shortlabels]{enumitem}
\usepackage{multirow}
\usepackage{amssymb}
\usepackage{bm}
\usepackage[accsupp]{axessibility}

\usepackage[breaklinks=true,bookmarks=false]{hyperref}

\iccvfinalcopy 

\def\iccvPaperID{2844} 

\ificcvfinal\fi

\begin{document}
\title{Continual Neural Mapping: \\Learning An Implicit Scene Representation from Sequential Observations}

\author{Zike Yan$^1$ \quad Yuxin Tian$^2$ \quad Xuesong Shi$^3$ \quad Ping Guo$^3$ \quad Peng Wang$^3$ \quad Hongbin Zha$^1$ \and
$^1$ Key Laboratory of Machine Perception (MOE), School of EECS, Peking University\\
PKU-SenseTime Machine Vision Joint Lab\\
$^2$ School of Automation Sceience and Electrical Engineering, Beihang University\\
$^3$ Intel Labs China\and
{\tt\small zike.yan@pku.edu.cn, tianyuxin@buaa.edu.cn,}\\ {\tt\small \{xuesong.shi, ping.guo, patricia.p.wang\}@intel.com, zha@cis.pku.edu.cn}
}

\twocolumn[{%
	\renewcommand\twocolumn[1][]{#1}%
	\maketitle
	\begin{center}
		\centering
		\includegraphics[width=0.79\linewidth]{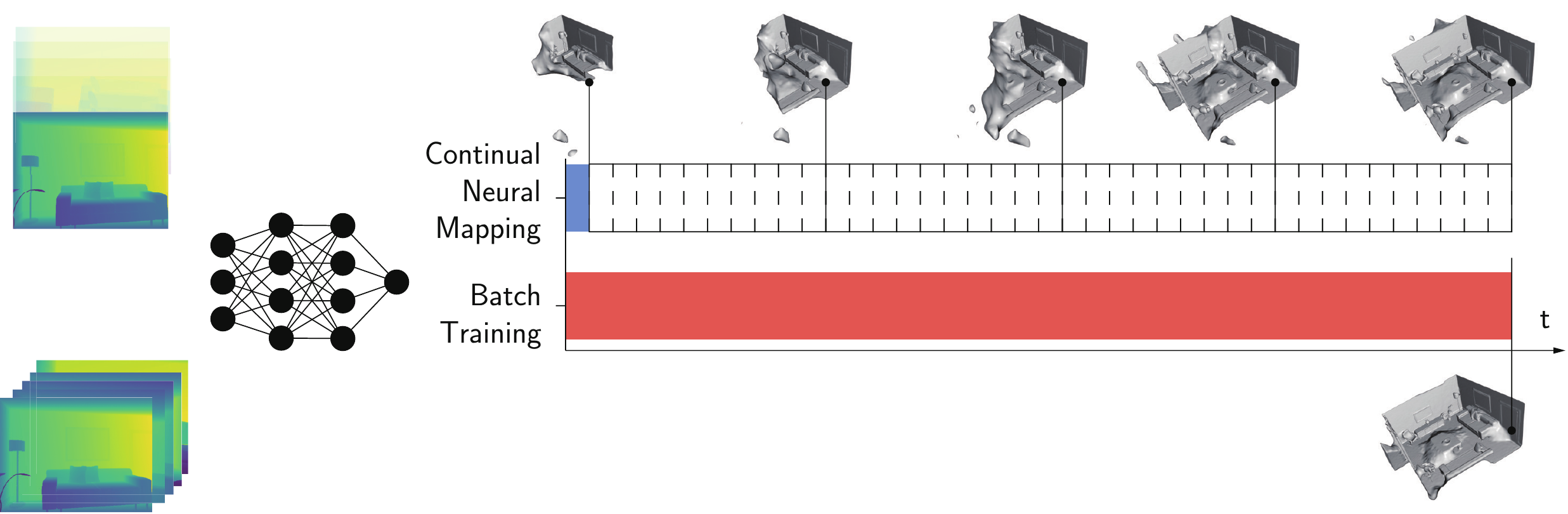}
		\captionof{figure}{Continual neural mapping learns scene properties from sequential data. By memorizing past experience within shared weights of a single network, we eliminate the need to store the entire sets of data or learn from scratch at each time.}
		\label{fig:cover}
	\end{center}%
}]

\newcommand*{\dictchar}[1]{
	\clearpage
	\twocolumn[
	\centerline{\parbox[c][4cm][c]{10cm}{%
			\centering
			\textbf{\Large
			{#1}}}}]
}

\newcommand{\beginsupplement}{%
	\setcounter{table}{0}
	\renewcommand{\thetable}{S\arabic{table}}%
	\setcounter{figure}{0}
	\renewcommand{\thefigure}{S\arabic{figure}}%
	\setcounter{section}{0}
	\renewcommand{\thesection}{\Roman{section}}%
}

\maketitle
\ificcvfinal\thispagestyle{empty}\fi

\begin{abstract}
   Recent advances have enabled a single neural network to serve as an implicit scene representation, establishing the mapping function between spatial coordinates and scene properties. In this paper, we make a further step towards continual learning of the implicit scene representation directly from sequential observations, namely Continual Neural Mapping. The proposed problem setting bridges the gap between batch-trained implicit neural representations and commonly used streaming data in robotics and vision communities. We introduce an experience replay approach to tackle an exemplary task of continual neural mapping: approximating a continuous signed distance function (SDF) from sequential depth images as a scene geometry representation. We show for the first time that a single network can represent scene geometry over time continually without catastrophic forgetting, while achieving promising trade-offs between accuracy and efficiency.
\end{abstract}

\section{Introduction}
\label{sec:intro}
Scene representations convert visual sensory data into compact forms. Recent trends~\cite{Sitzmann2020nips, Mildenhall2020eccv} show that the mapping function $\bm{y} = f(\bm{x};\theta)$ between the spatial coordinate $\bm{x}$ and the scene property $\bm{y}$ can serve as an implicit scene representation parameterized by a single neural network $\theta$. Such a new paradigm has drawn significant attention: the neural network defined in a continuous and differentiable function space can be trained to recover fine-grained details at scene scale with efficient memory consumption, which offers great benefits over alternatives.

However, batch training of the implicit neural representation is impractical and inefficient when dealing with possibly unending streams of data. To handle the sequential observations and obtain a globally consistent representation over time, conventional approaches turn to a data fusion paradigm. A discretized scene representation is pre-defined in memory-inefficient parameter space and updated according to perceived observations at each time. The gap between the emerging neural representation paradigm and the conventional data fusion paradigm addresses a critical issue: \emph{how we can learn an implicit neural representation continually from sequential observations}?
 
In this paper, we introduce a novel problem setting of \emph{continual neural mapping}. The central idea is to maintain a continually updated neural network at each time to approximate the mapping function $f(\cdot)$ within the environment. Past observations $(\bm{x}^{1:t}, \bm{y}^{1:t})$ are marginalized out and summarized into compact neural network parameters $\theta^t$ during training. The neural network not only serves as a memory of sequential data, but also makes predictions of scene properties within the entire environment. The prediction-updating fashion leads to a self-improved mapping function when constantly exploring the environment, which resembles human-like learning scenarios from a continual learning perspective.

We instantiate the proposed continual neural mapping problem by tackling the SDF approximation from sequential depth images. We propose an experience replay approach that distills past experience to guide the prediction without catastrophic forgetting. Experimental results demonstrate that the proposed method outperforms batch re-training/fine-tuning baselines and obtains comparable results against state-of-the-art approaches. The key contributions of our work are summarized as follows:

- We are the first to address the problem of learning an implicit neural scene representation continually from sequential data, namely \emph{continual neural mapping};

- We deal with the problem of SDF approximation from sequential data under the proposed continual neural mapping setting, outperforming competitive approaches;

- We propose an experience replay method to learn scene geometry continually without catastrophic forgetting. The memory consumption and training time are orders of magnitude less than the batch re-training baseline.

\section{Related Work}
\label{sec:relatedwork}
The proposed continual neural mapping setting lies in the intersection of implicit neural representation, 3D data fusion, and continual learning. In this section, we review the most related work in each area and highlight the major differences over the proposed problem setting.

\subsection{Implicit Neural Representation}
Implicit neural representation takes a neural network as the continuous mapping function between the spatial coordinates and the scene properties. Shape-conditioned representations concatenate the coordinate $\bm{x} $ and a latent shape embedding $\bm{z}$ to represent multiple shape instances as $\bm{y}=f(\bm{x}, \bm{z};\theta)$. The shape embedding $\bm{z}$ is latter conducted in a local fashion to recover fine-grained details at scene scale~\cite{Jiang2020cvpr, Peng2020eccv, Chabra2020eccv}. The output properties $\bm{y}$ infered from the shape-conditioned representations vary across shape~\cite{Park2019cvpr, Mescheder2019cvpr, Chen2019cvpr}, appearance~\cite{Niemeyer2020cvpr, Oechsle2020_3dv, Schwarz2020nips}, and motion~\cite{Niemeyer2019ICCV}.  Another line uses neural networks to regress the parameters of decomposed primitives directly from the input point set as $\{\bm{m}_j\} = f(\{\bm{x}_i\}; \theta)$. The regressed primitive parameters are then grouped together as the entire shape parameter space, representing the scene geometry as $\bm{y} = \phi(\{\bm{m}_j\}, \bm{x})$. Commonly used primitives include hyperplanes~\cite{Deng2020cvpr}, Gaussian mixtures~\cite{Hertz2020cvpr, Genova2020cvpr, Genova2019iccv}, volumes~\cite{Tulsiani2017CVPR}, and local planes~\cite{Chen2020cvpr}. 

Recent work has investigated the mapping from spatial coordinates to scene properties directly through MLPs as $\bm{y}=f(\bm{x};\theta)$. The high-frequency details can be preserved well with the help of positional encoding~\cite{Mildenhall2020eccv}, Fourier feature mapping~\cite{Tancik2020nips}, or the periodic activation~\cite{Sitzmann2020nips}. We extend the implicit neural representation to a continual learning fashion, where an implicit representation can be directly learned from sequential data without computationally expensive re-training or catastrophic forgetting.

\subsection{Incremental Depth Fusion}
Conventional depth fusion paradigm aims to maintain a pre-defined output representation instead of the implicit network parameters. The mapping between coordinates and the output representation is accomplished through a deterministic data assignment, where the parameters of the output representation are incrementally updated through weighted averaging according to streaming observations. Most commonly used representations for incremental depth fusion include volumetric TSDF~\cite{Curless1996siggraph, Kinectfusion, VoxelHashing, Whelan2015ijrr, Dai2017tog} and surfel~\cite{surfels, Whelan2015rss, Keller2013_3dv, Schops2019cvpr, SurfelMeshing}. On account that the representation is defined in the continuous output range, discretization is inevitable. Postprocessing is usually conducted to transform the discrete representation into a watertight mesh~\cite{marchingcubes} or render the scene as a view-dependent dense image.

Recent advances have fostered a learning-based depth fusion fashion. \cite{Bloesch2018cvpr, Zhi2019cvpr, Czarnowski2020ral} seek to find a compact and optimisable feature for estimated monocular depth, where the entire scene is represented by a set of keyframes with low-dimensional depth codes. Further extensions utilize neural networks to learn aggregation of learned image/depth features in a latent space~\cite{Riegler2017_3dv, Paschalidou2018cvpr, Ji2017iccv, Xie2019iccv, Murez2020eccv, Weder2021cvpr} as the global scene representation. RoutedFusion~\cite{Weder2020cvpr} follows the conventional TSDF fusion pipeline and learns how volumetric TSDF is updated. However, all previous approaches view depth fusion as a deterministic learnable operation. We, on the other hand, address the problem of knowledge fusion as a continual learning (training) procedure, where the neural network serves as a self-improving scene representation with parameters updated continually.
\begin{figure*}[t]
	\centering
	\subfloat[Multi-task learning]{
		\begin{minipage}{0.25\linewidth}
			\includegraphics[width=0.95\linewidth]{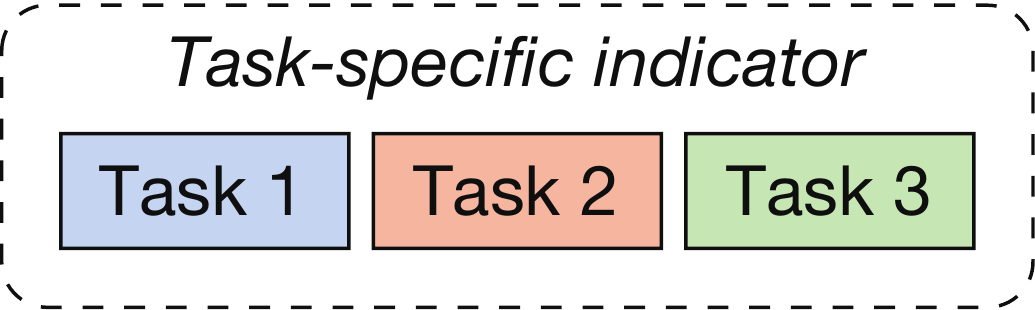}
		\end{minipage}
	}
	\subfloat[Fine-tuning]{
		\begin{minipage}{0.25\linewidth}
			\includegraphics[width=0.95\linewidth]{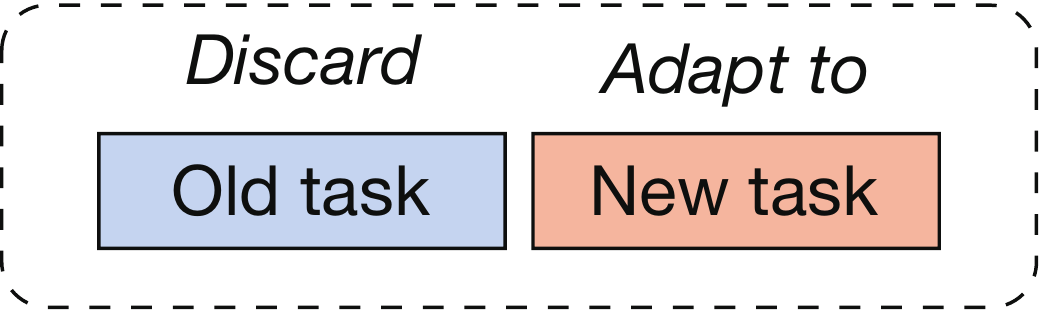}
		\end{minipage}
	}
	\subfloat[Batch re-training]{
		\begin{minipage}{0.25\linewidth}
			\includegraphics[width=0.95\linewidth]{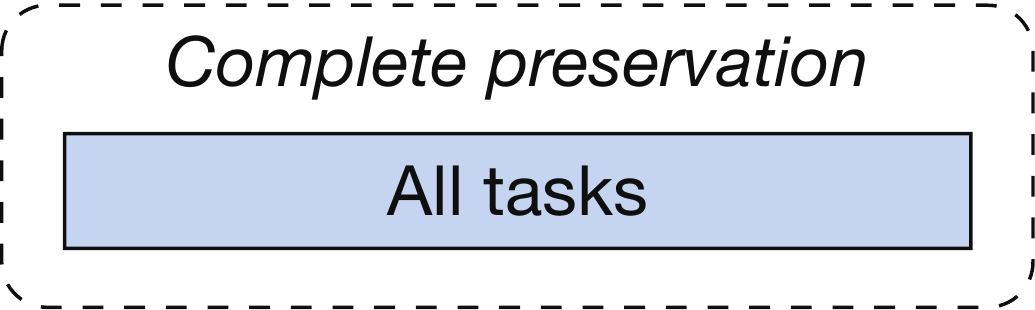}
		\end{minipage}
	}
	\subfloat[Continual learning]{
		\begin{minipage}{0.25\linewidth}
			\includegraphics[width=0.95\linewidth]{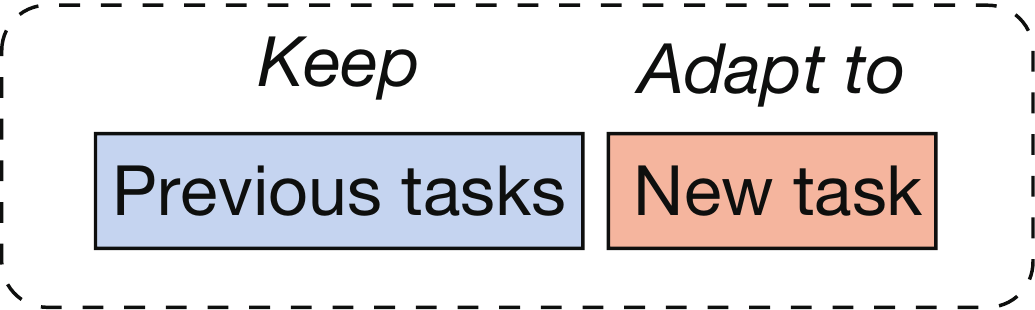}
		\end{minipage}
	}
	\caption{Relevant learning paradigms.}
	\label{fig:learningParadigm}
\end{figure*}

\subsection{Continual Learning}
The proposed \emph{continual neural mapping} problem targets the updating of network parameters at each time when new observations arrive that lead to a self-improved mapping function. This problem setting falls into a continual learning~\cite{Parisi2019nn} category: the streaming data are no longer iid-sampled but highly correlated to adjacent ones. We want to update the mapping function in the newly observed areas while preserving an accurate mapping in previously visited regions without forgetting. 

Following~\cite{Delange2021pami}, existing methods can be generally categorized into three kinds: regularization, parameter isolation, and replay.  Regularization-based methods aim to enforce parameter consistency during the training process by penalizing the changes of important parameters~\cite{Kirkpatrick2017ewc, Zenke2017icml, Chaudhry2018eccv, Aljundi2018eccv} or through knowledge distallation~\cite{Li2017pami, Rannen2017iccv}. On the other hand, parameter-isolation methods assume that different subsets of the network parameters are attributed to different tasks, thus leading to a flexible gating mechanism. Isolated parameters for each sub-network can be achieved through a dynamically expanded network~\cite{Rusu2016arXiv, Lee2020iclr, Aljundi2017cvpr, Yoon2018iclr, Rao2019nips} or through task-specific masking~\cite{Mallya2018cvpr, Mallya2018eccv, Serra2018icml}. Finally, replay-based methods store samples or generate pseudo samples as the memory of old knowledge. Special attention is devoted to different sample selection strategies~\cite{Rebuffi2017cvpr, Rolnick2019nips}, sample generation strategies~\cite{Atkinson2021nn, Rios2019ijcai, Xiang2019iccv}, and optimization constraints~\cite{Lopez2017nips, Chaudhry2018iclr}. In this paper, we propose an experience replay approach to tackle the proposed continual neural mapping problem by leveraging past experience to guide the continual learning of new observations.

\section{Continual Neural Mapping}
\label{sec:overview}
In this section, we formalize the proposed continual neural mapping problem setting. The connections to relevant learning paradigms are clarified afterwards.
\subsection{Problem Statement}
We consider a general setting within a 3D environment $\mathcal{W}$, where sequential data $\mathcal{D}^t$ are constantly captured. The data $\mathcal{D}^t=\{(\bm{x}_i^t, \bm{y}_i^t)\}_{n^t}$ consist of $n^t$ tuples of spatial coordinates $\bm{x}_i^t \in \Omega^t$ and the corresponding scene properties $\bm{y}_i^t$ with observed areas $\Omega^t \subset \mathcal{W}$ specified. The objective of the continual neural mapping is to learn a mapping function $f(\cdot)$ parameterized by a neural network $\theta^t$ continually from the observed data $\mathcal{D}^t$ to depict the connections between the spatial coordinates and the scene properties as:
\begin{equation}
	\label{eq:mapping}
	\bm{y} = f(\bm{x};\theta^t), \forall \bm{x} \in \mathcal{W}.
\end{equation}

\noindent\textbf{Knowledge transfer.} The mapping function $f(\cdot)$ serves as an implicit neural representation for the 3D environment, which can be queried at any time to predict the scene property $\bm{y}$ given the spatial coordinate $\bm{x}$. For previously visited areas $\bm{x}\in \Omega^{1:t}$, the mapping function serves as a compact memory of past observations $\mathcal{D}^{1:t}$. This is related to \emph{backward transfer}~\cite{Lesort2020if, Lopez2017nips}, where the neural network not only memorizes existing data, but also leads to better performance on previously visited areas when learning from new observations. On the other hand, for unseen areas $\bm{x} \in \mathcal{W} \cap \overline{\Omega^{1:t}}$, the mapping function serves as a predictor. \emph{Forward transfer} may be facilitated that distills knowledge and skills for future exploration. Consequently, continual neural mapping alleviates the need for storing the entire dataset $\mathcal{D}^{1:t}$ while preserving the complete mapping function within the environment, guaranteeing a quick convergence to new observations.

\noindent\textbf{Challenges.} 
The objectiveness of the proposed continual neural mapping problem is to find an optimal neural network that shares parameters $\theta^t$ across all previous tasks\footnote{\emph{Task} refers to a particular period of time where the data distribution is stationary and the objective function is constant~\cite{Lesort2020if}.} as:
\begin{equation}
	\label{eq:overall_objectiveness} 
    \arg\min_{\theta_t}\frac{1}{\|\mathcal{D}^{1:t}\|}\sum_{\mathcal{D}^{1:t}} \mathcal{L}(f(\bm{x};\theta^t), \bm{y})
\end{equation}

The major challenge lies in the gap between the proposed problem setting and the conventional Empirical Risk Minimization (ERM) principle~\cite{Vapnik1992ERM}: the streaming data $\mathcal{D}^t$ lead to constant distribution shift. As non-stationary data distribution breaks the iid-sampled assumption, a learning solution is required to model the overall distribution of past observations without the need to store the entire dataset $\mathcal{D}^{1:t}$.

\begin{figure}[!t]
	\centering
	\includegraphics[width=0.9\linewidth]{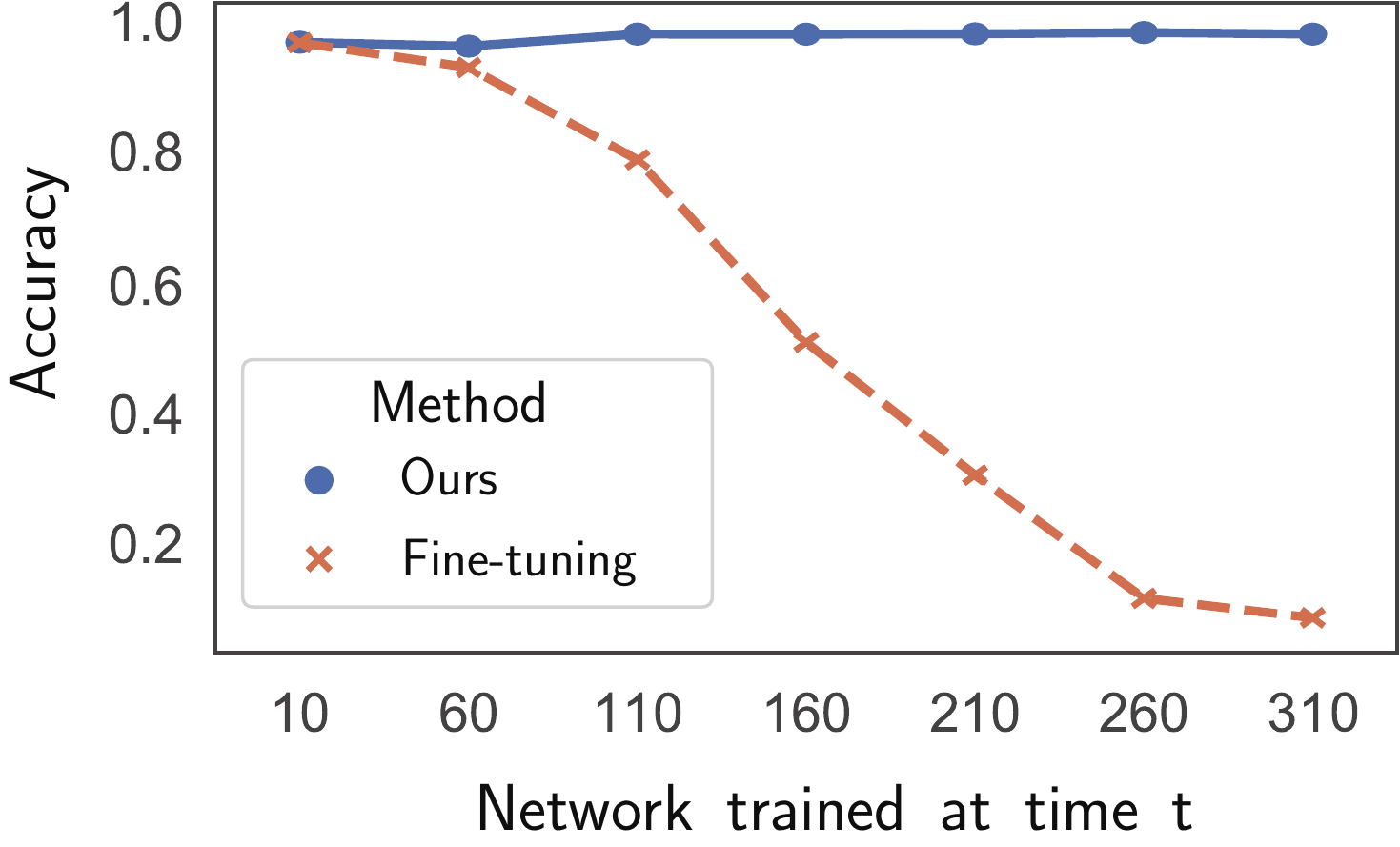}
	\captionof{figure}{We validate the issue of forgetting by estimating per-point SDF value for the first frame with current network parameters. The percentage of $|f(\bm{x};\theta^t)|<0.01$ declines drastically for the fine-tuning baseline, while the proposed method maintains a consistent accuracy level across time.}
	\label{fig:forget}
\end{figure}


\subsection{Connections to Relevant Learning Paradigms}
Though the objectiveness in (\ref{eq:overall_objectiveness}) is a joint optimization of temporally seperated tasks $\mathcal{L}(f(\bm{x};\theta), \bm{y}), (\bm{x}, \bm{y})\in\mathcal{D}^{t}$, continual neural mapping can be understood from the perspectives of four relevant learning paradigms as illustrated in Fig.~\ref{fig:learningParadigm}. Clarifying our problem setting from the most relevant continual learning perspective, continual neural mapping falls into a \emph{domain-incremental}~\cite{Van2019arxiv} continual learning scenario: we aim to maintain a globally consistent representation with a single network from sequential data, where data distribution shifts and the objective remains the same.

On the other hand, multi-task learning splits the training process into a set of dependent tasks and optimizes all tasks jointly. For our continual neural mapping setting, the task boundary is unknown, thus requiring a continual task identifier that assigns training data to specific tasks consistently over time. Meanwhile, backward transfer addresses the problem of constant network adaptation for all tasks, which is opposed to conventional multi-task learning that fixes the network once the model is deployed~\cite{Delange2021pami}. The fine-tuning strategy maintains a single network consecutively, where network parameters of a new task are initialized with that of the last task. However, as neural networks tend to be overly plastic~\cite{Mirzadeh2020nips} from the "plasticity-stability dilemma" perspective~\cite{Mermillod2013dilemma}, the performance of early tasks will degrade on current network parameters (Fig.~\ref{fig:forget}), namely \emph{catastrophic forgetting}~\cite{Mccloskey1989catastrophic}. Finally, batch re-training preserves all previously observed data $\bm{x}^{1:t}$ to satisfy the iid-sampled assumption. However, batch re-training learns a new model at each time from scratch without exploiting past experience. The linearly-growing number of training data results in expensive memory consumption and computational cost.
\section{Example: SDF Regression}
\label{sec:algorithm}
In this section, we instantiate the proposed continual neural mapping on the task of scene geometry approximation. The objective is a special case of (\ref{eq:overall_objectiveness}) that defines the mapping function $f(\cdot)$ as the SDF parameterized by a single multilayer perceptron (MLP), representing the 3D surface as a zero level-set $\mathcal{M}$:
\begin{equation}
	\label{eq:level_set} 
	\mathcal{M} = \{\bm{x}\in\mathbb{R}^3|f(\bm{x};\theta^t)=0\}, f(\cdot):\mathbb{R}^3 \mapsto \mathbb{R}.
\end{equation}

In a batch training setting, the problem is studied by~\cite{Gropp2020icml, Sitzmann2020nips} and solved as an Eikonal boundary value problem. The continuous SDF can be fit from oriented point cloud data that are iid-sampled from closed surfaces. We take a step further to tackle a more realistic and challenging case that continually learns SDFs parameterized by a single MLP from streaming posed depth images.

\subsection{Solution}
In practice, we split the energy function (\ref{eq:overall_objectiveness}) into two terms with equal weights as:
\begin{equation}
	\label{eq:split_terms} 
    \sum_{\mathcal{D}^{1:t-1}} \mathcal{L}(f(\bm{x};\theta^t), \bm{y}) +\sum_{\mathcal{D}^{t}} \mathcal{L}(f(\bm{x};\theta^t), \bm{y}),
\end{equation}
where the loss function $\mathcal{L}(f(\bm{x};\theta^t), \bm{y})$ consists of a data term $|f(\bm{x};\theta)|$, an Eikonal term $|\|\nabla_{\bm{x}}f(\bm{x};\theta)\|-1|$, a normal constraint $|\nabla_{\bm{x}}f(\bm{x};\theta)-\bm{n}|$, and an off-surface constraint $\psi(f(\bm{x};\theta))=\exp(-\alpha\cdot|f(\bm{x};\theta)|), \alpha \gg 1$ following~\cite{Sitzmann2020nips}.

\begin{figure}[!t]
	\centering
	\includegraphics[width=0.83\linewidth]{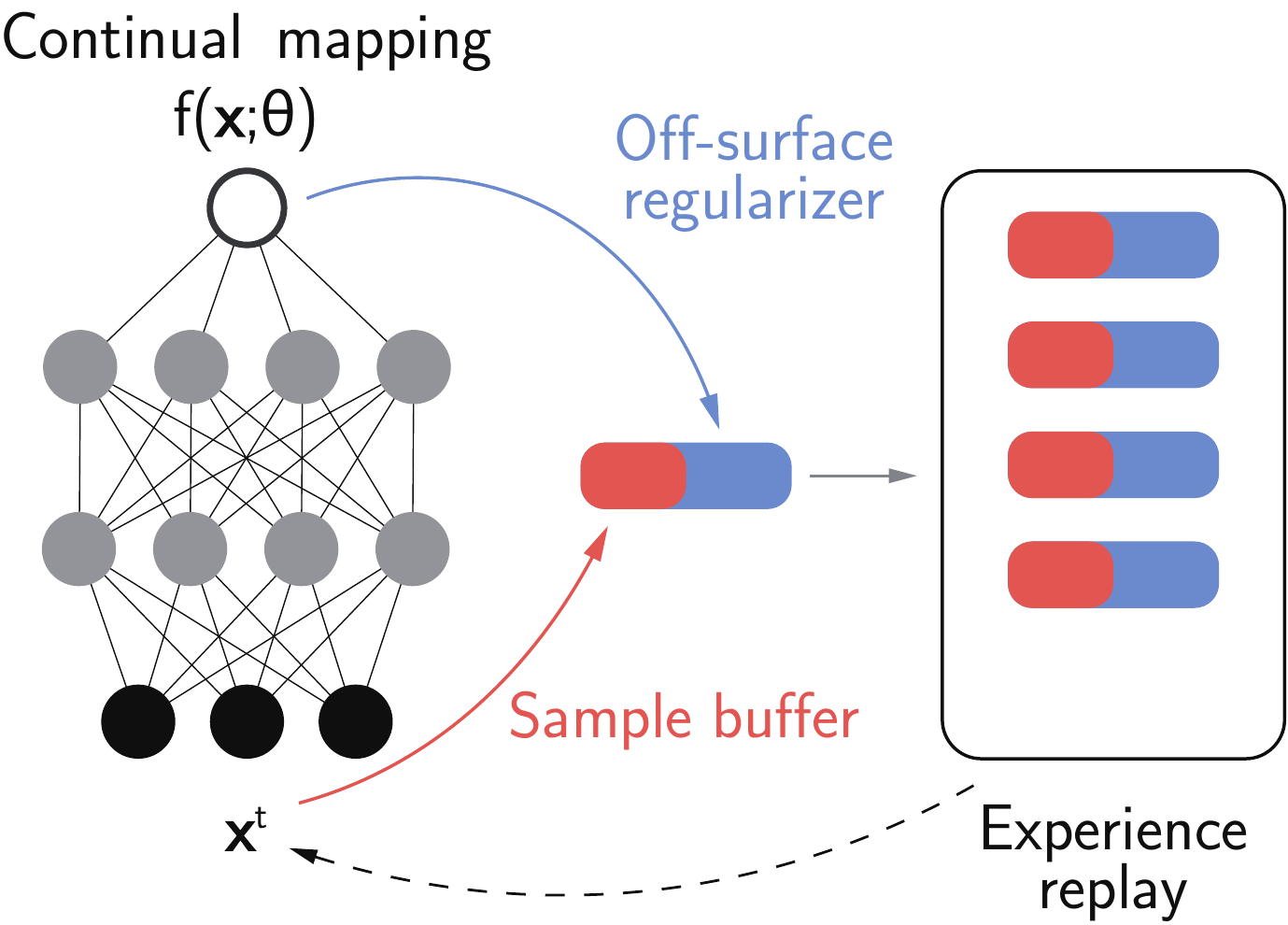}
	\captionof{figure}{During training, newly observed data are preserved within a fixed size of buffer to constrain zero level-set. Off-surface samples are guided by last network parameters to penalize sign deviation.}
	\label{fig:replay}
\end{figure}

The split terms in (\ref{eq:split_terms}) can be understood as a combinatory constraint of the current observation $\mathcal{D}^t$ and the past experience $\mathcal{D}^{1:t-1}$. Notice that in the proposed continual neural mapping setting, the entire sequence of raw data $\mathcal{D}^{1:t-1}$ should not be preserved, we resort to an experience replay method to model the past experience as illustrated in Fig.~\ref{fig:replay}.
 
\subsection{Experience Replay}
\label{subsec:experience_replay}
As mentioned in Sec.~\ref{sec:intro}, the maintained neural network serves as a compact memory of previous observations. Hence, random coordinates with SDF values approximated by the last neural network $\{\bm{x}_i,f(\bm{x}_i;\theta^{t-1})\}_{n^t}$ can be viewed as iid samples replayed from past experience to regularize the current network. The replayed samples are twofold: 1) off-surface samples to regularize the distance sign; 2) zero level-set samples to regularize the data term and the normal constraint. 
\begin{figure}
	\begin{minipage}{0.25\linewidth}
		\begin{minipage}[b]{0.25\linewidth}
			\centering
			\captionsetup{type=table}
			\begin{tabular}{lccc}
				\hline
				 & 1 & 2 & 3 \\
				\hline
			    Camera a & + & -& -\\
				Camera b & - & + & -\\
				Actual & + & + & +\\
				\hline
			\end{tabular}
		\end{minipage}\vspace{0.03\textwidth}\\
	\end{minipage}
    \hspace{0.65in}
    \begin{minipage}{0.6\linewidth}
	    \includegraphics[width=0.9\linewidth]{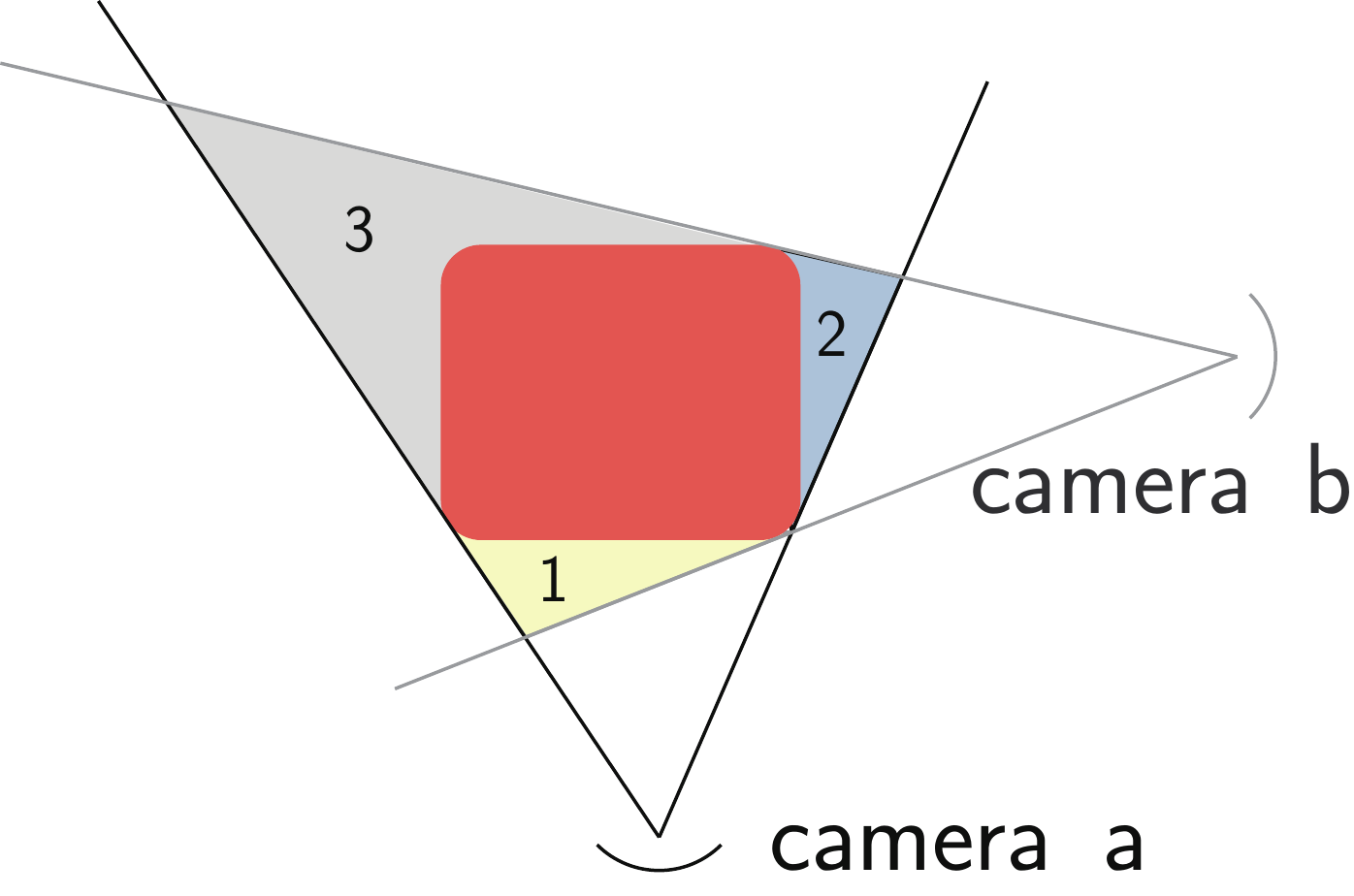}
    \end{minipage}
	\caption{It is obvious that sign reasoning for areas behind the red surface may be false negative. We need to incorporate previous knowledge to better regularize the sign of off-surface samples.}
	\label{fig:sign}
\end{figure}

For off-surface samples $\{\bm{x}_{o}, f(\bm{x}_{o}, \theta^{t-1})\}$, we backproject the points to the camera coordinate at time $t$ and compare them against the depth map. As illustrated in Fig.~\ref{fig:sign}, points within the frustum that are in front of the surface should be true positive, while points within the frunstum that are behind the surface may be false negatives. Hence, we label samples with positive SDF approximation $f(\bm{x}_o;\theta^{t-1})>0$ or the samples that fall in front of the surface within the frustum as positive, and label samples that fall behind the surface within the frustum with negative SDF approximation $f(\bm{x}_o;\theta^{t-1})<0$ as negative. The sign regularizer for off-surface samples is then defined as:
\begin{equation}
	\label{eq:sign_regularizer} 
    \psi_{s}(f(\bm{x};\theta))=
	\begin{cases}
		\exp(-\alpha \cdot f(\bm{x};\theta))& \text{if positive}\\
		\exp(\alpha \cdot f(\bm{x};\theta))& \text{if negative}
	\end{cases}
\end{equation}

For zero level-set samples $\bm{x}_z$, one intuitive solution is to construct voxel grids and estimate the SDF value for each vertex. The zero level-set samples can then be easily obtained on the extracted triangle mesh. However, due to incomplete observations, spurious zero level-set samples may be generated in unseen areas. Additional maintenance of occupancy status for each voxel grid is required to eliminate erroneous samples, which is memory inefficient. Here, we take a simple solution to randomly down-sample previous observations and maintain a fixed size of buffer data~\cite{Lee2020iclr} to regularize the data term $|f(\bm{x}_z;\theta)|$. Experimental results show that this simple solution is effective enough to store previous knowledge without catastrophic forgetting. 
\section{Experiments}
In this section, we demonstrate that the proposed continual neural mapping setting succeeds in representing constantly observed scene geometry with a single neural network from scratch. The recovered accuracy is comparable against competitive methods with orders of magnitude less memory consumption.

\subsection{Experimental Setup}
The experiments were conducted on a desktop PC with an Intel Core i7-8700 (12 cores @ 3.2 GHz), 32GB of RAM, and a single NVIDIA GeForce RTX 2080Ti. 

\noindent\textbf{Model.} We use a single 5-layer SIREN MLP~\cite{Sitzmann2020nips} with $256$ units in each layer as our base network model.

\noindent\textbf{Baselines.} Following~\cite{Delange2021pami, Lee2020iclr}, the \emph{fine-tuning} baseline is initialized with the last model parameters at each time and is naively trained for each frame; the \emph{re-training} baseline is trained with the entire sequence of data following the signed distance function setup of SIREN~\cite{Sitzmann2020nips}. To further study the effect of the replay buffer, we also provide a \emph{re-initialization} baseline that learns from scratch with the buffer data for each frame. Adam optimizer is adopted with a learning rate of $0.0001$. The data for each method are trained for $1500$ epochs if not specified, while the first frame is trained for $10000$ epochs to ensure nice initialization. 

\noindent\textbf{Dataset.} We mainly evaluate our results quantitatively and qualitatively on the synthetic ICL-NUIM livingroom dataset~\cite{ICL}. Additional qualitative evaluation is conducted on the real TUM dataset~\cite{TUM}. The entire sequence is downsampled due to the extremely high cost of the batch re-training baseline. The normal is estimated and oriented towards the camera location using Open3D~\cite{Open3D}.

\subsection{Model Analysis}
\label{subsec:analysis}
We provide an in-depth analysis of the proposed experience replay method by comparing it with the aforementioned baselines and state-of-the-art methods. We refer readers to our supplementary video for better visualizing the continual changes over time.

\begin{figure*}[btp]
	\centering
	\subfloat[{Batch re-training} ($0.002/0.003$)]{
		\begin{minipage}[t]{0.3\linewidth}
			\centering
			\includegraphics[width=0.98\linewidth]{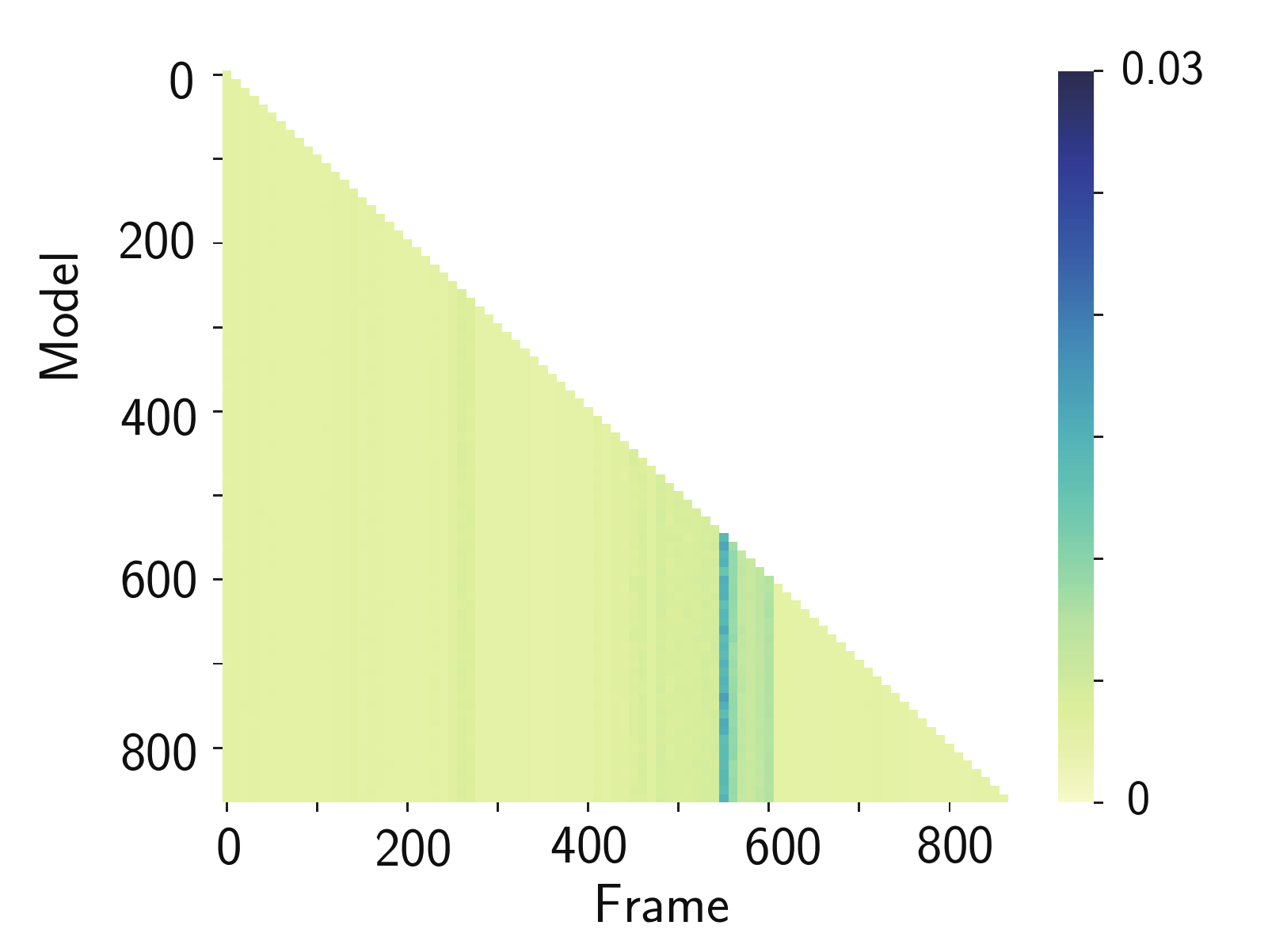}
		\end{minipage}
	}
	\subfloat[{Fine-tuning} ($0.496/0.627$)]{
		\begin{minipage}[t]{0.3\linewidth}
			\centering
			\includegraphics[width=0.98\linewidth]{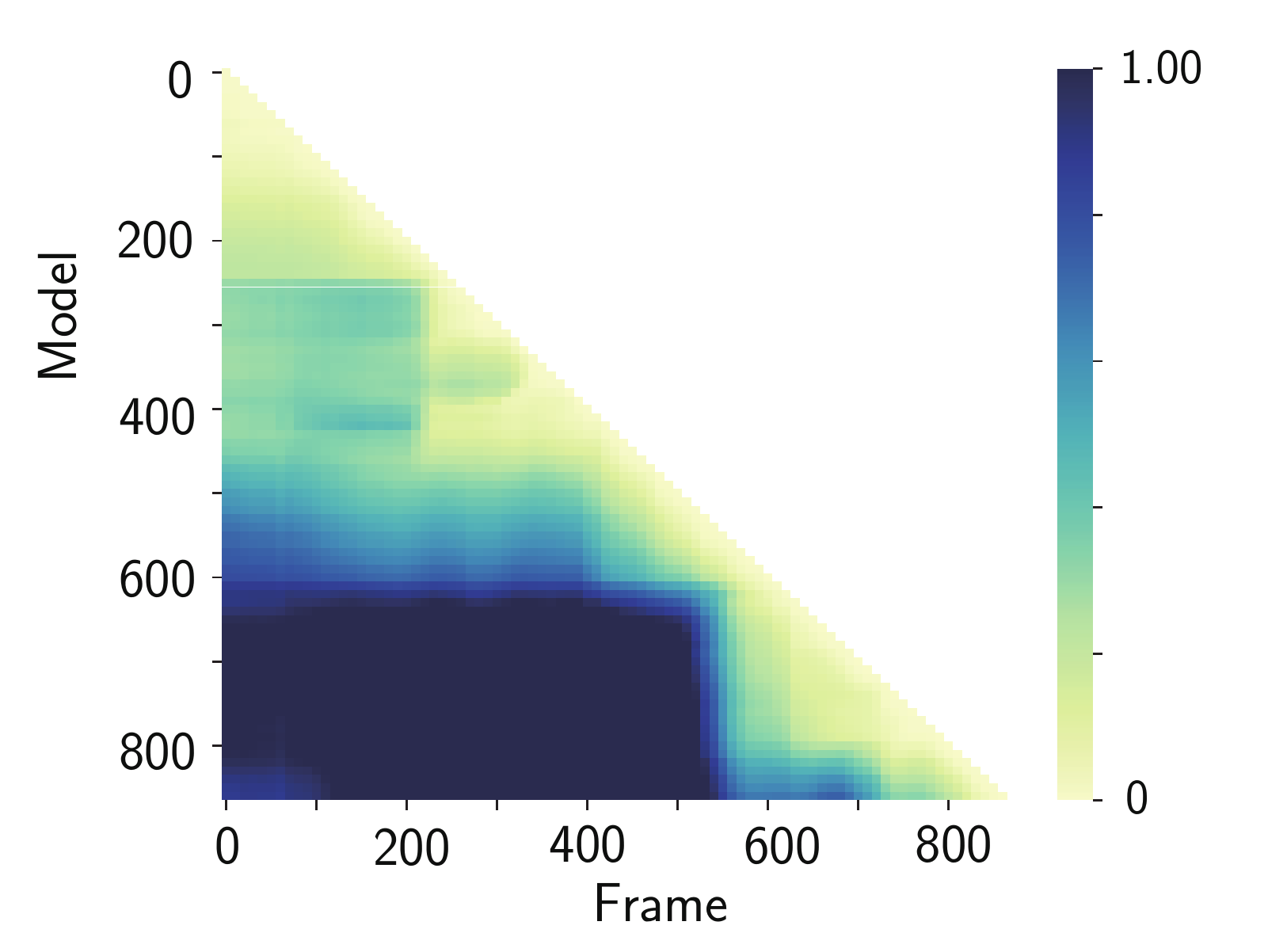}
		\end{minipage}
	}
	\subfloat[Ours ($0.003/0.004$)]{
		\begin{minipage}[t]{0.3\linewidth}
			\centering
			\includegraphics[width=0.98\linewidth]{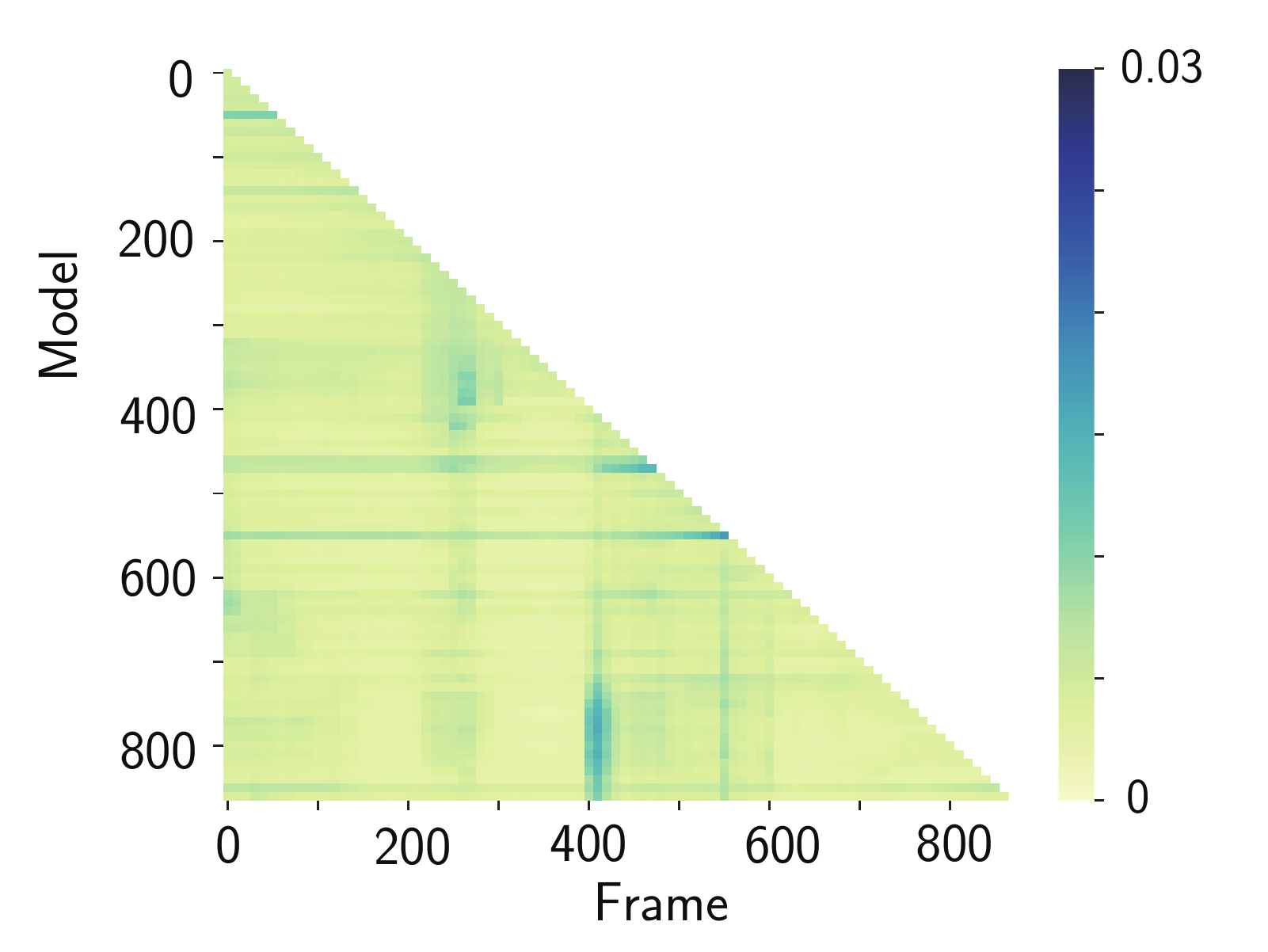}
		\end{minipage}
	}
	\caption{The accuracy heatmap with overall mean/std.~(m) of each method on the ICL dataset. The heatmap value at (m,n) is the mean SDF approximation of all points from frame $m$ using $n$th network parameters. Noticeably, the proposed method maintains consistent accuracy for all frames, while the fine-tuning baseline suffers from catastrophic forgetting severely.}
	\label{fig:heatmap}
\end{figure*}

\noindent\textbf{Continual neural mapping without forgetting.} We first assess the forgetting issue under the proposed continual neural mapping setting. The objective is to establish an accurate mapping between spatial coordinates and the corresponding SDF value in previously visited areas. Notice that the depth data sampled from the synthetic ICL dataset are exactly surface samples with zero distance. We calculate the mean distance of each frame $\bm{x}^t$ using the learned network parameters $\theta^t$ at each time. As illustrated in Fig.~\ref{fig:heatmap}, the proposed method achieves comparable accuracy against the batch re-training baseline while eliminating the catastrophic forgetting issue compared to the fine-tuning baseline. 

\begin{figure*}[tb!]
	\centering
	\includegraphics[width=0.97\linewidth]{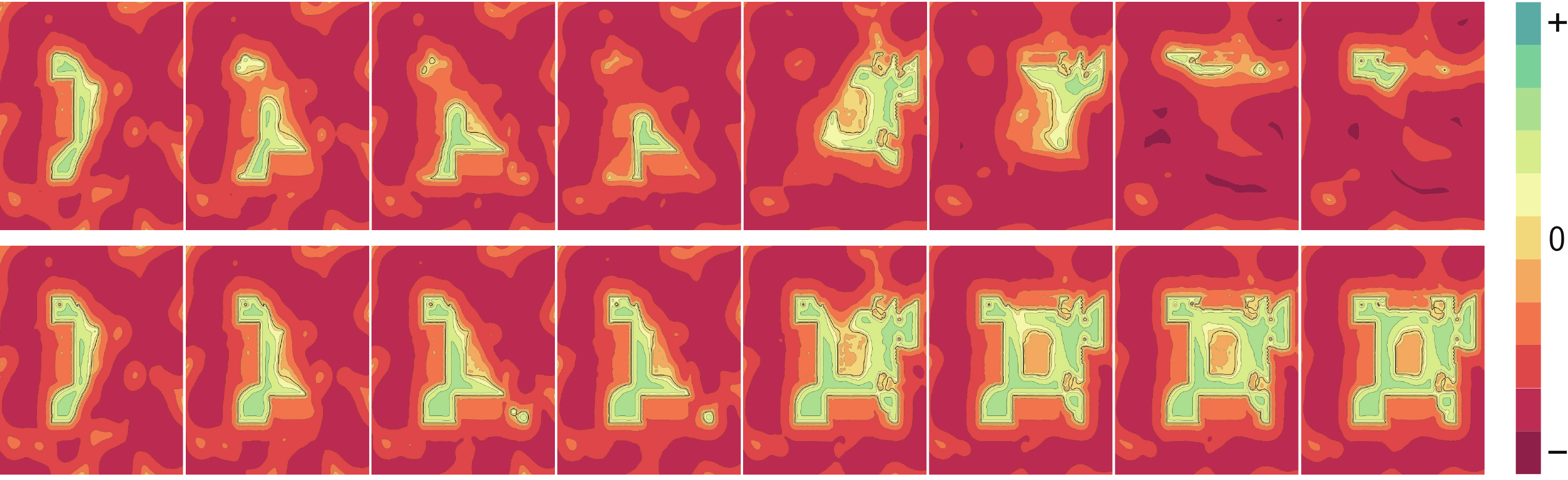}
	\caption{Top view visualization on the ICL dataset of SDF approximation from frame $100$ to $800$. Fine-tuning baseline (top) suffers from catastrophic forgetting, while ours (bottom) recover the scene geometry continually from sequential data.}
	\label{fig:sdf_vis}
\end{figure*}

A similar conclusion can be drawn from the 2D visualization of the SDF approximation. As illustrated in Fig.~\ref{fig:sdf_vis}, the fine-tuning baseline quickly forgets the geometry of previously visited areas, while the proposed method maintains a gradually improved SDF approximation during the exploration of the mobile sensor.

\noindent\textbf{Effective solution of experience replay.} We further study the role of experience replay in our continual neural mapping setup. Revisiting Sec.~\ref{subsec:experience_replay}, past experience is used to initialize the network weight, regularize the sign of free space, and constrain the zero level-set. We find that these three issues are essential to the problem of SDF regression from sequential data, guaranteeing past knowledge transfer for accurate SDF approximation.

\begin{figure}[!bth]
	\centering
	\includegraphics[width=0.9\linewidth]{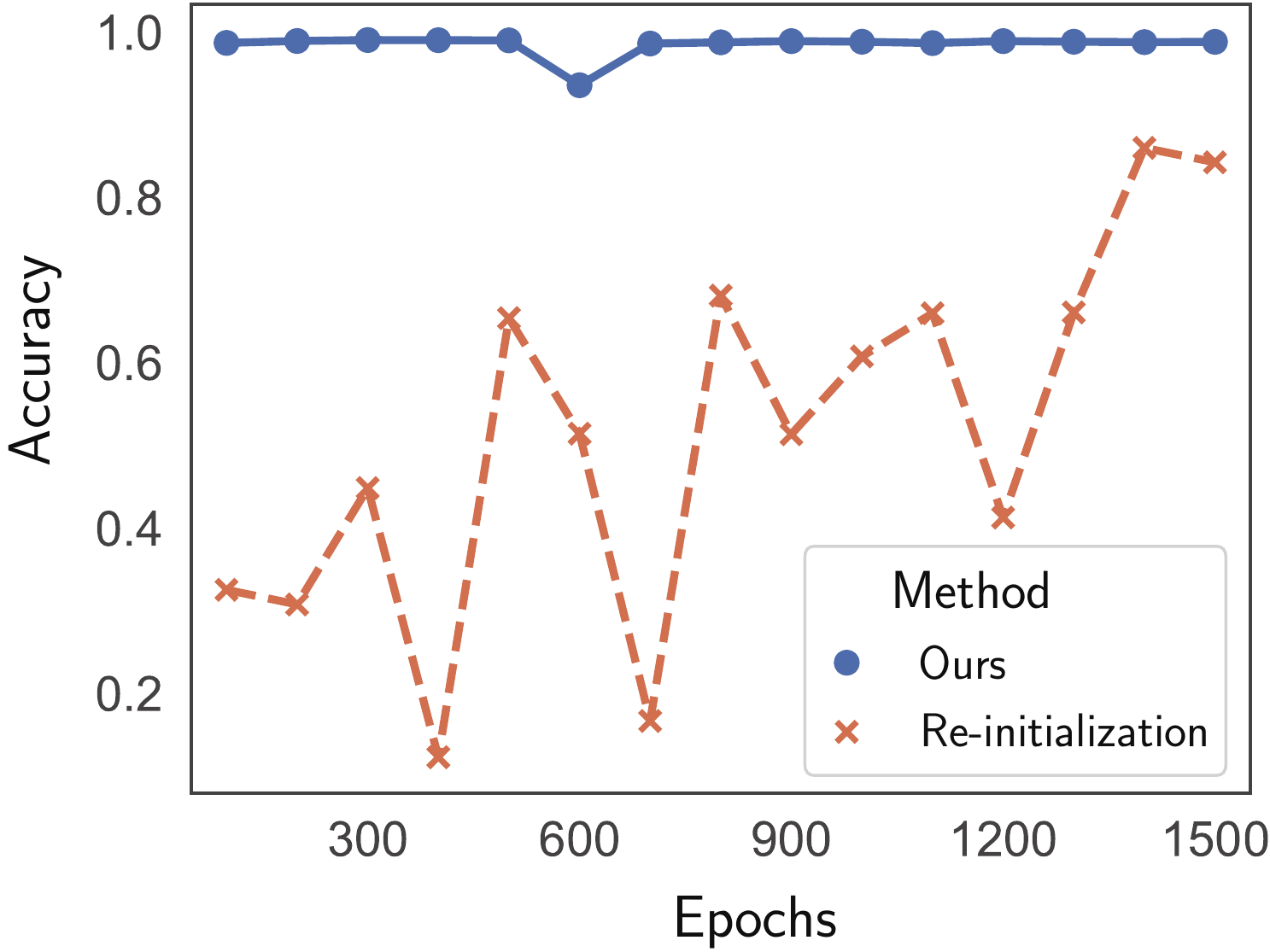}
	\caption{Compared to re-initialization, parameter sharing between frames is beneficial for knowledge distallation, resulting in faster convergence with better performance.}
	\label{fig:reinit_loss}
\end{figure}

\begin{figure*}[htbp]
	\centering
	\subfloat[Re-initialization ($1500$ epochs)]{
		\begin{minipage}{0.23\linewidth}
			\centering
			\includegraphics[width=0.95\linewidth]{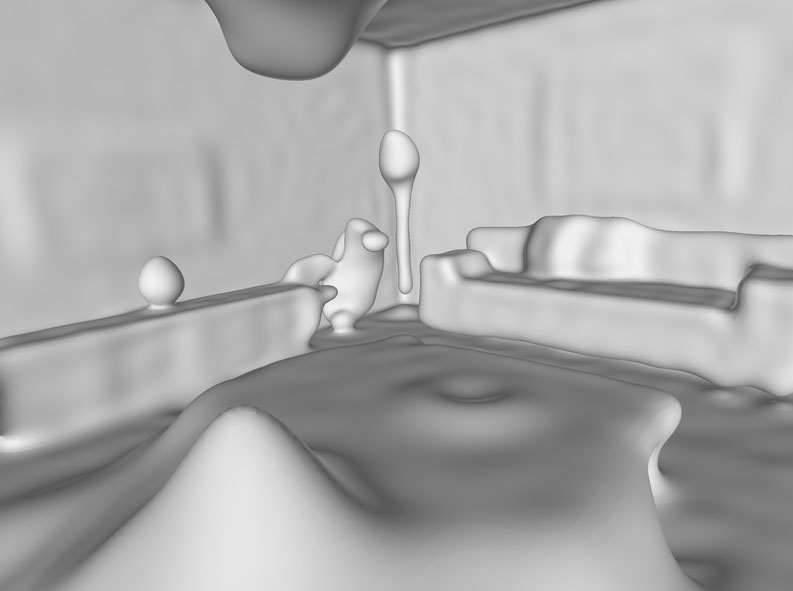}\\
			\includegraphics[width=0.95\linewidth]{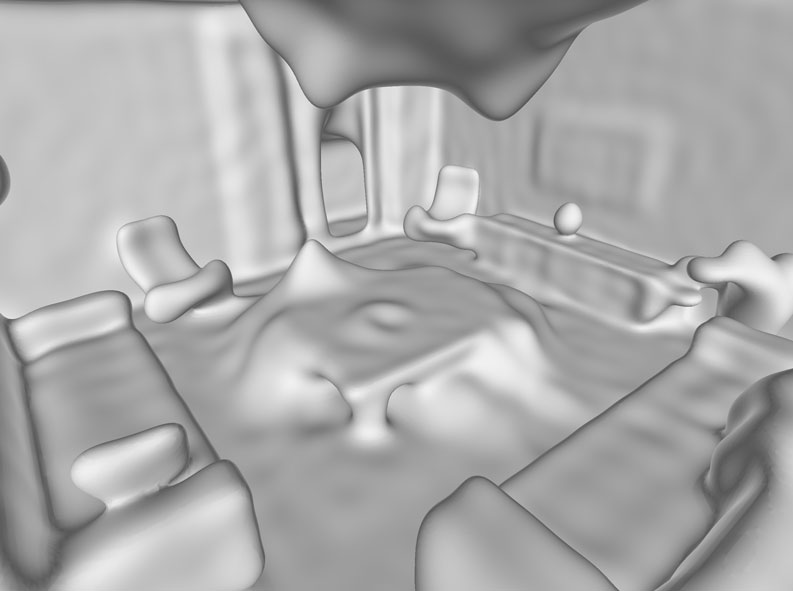}\\
			\includegraphics[width=0.95\linewidth]{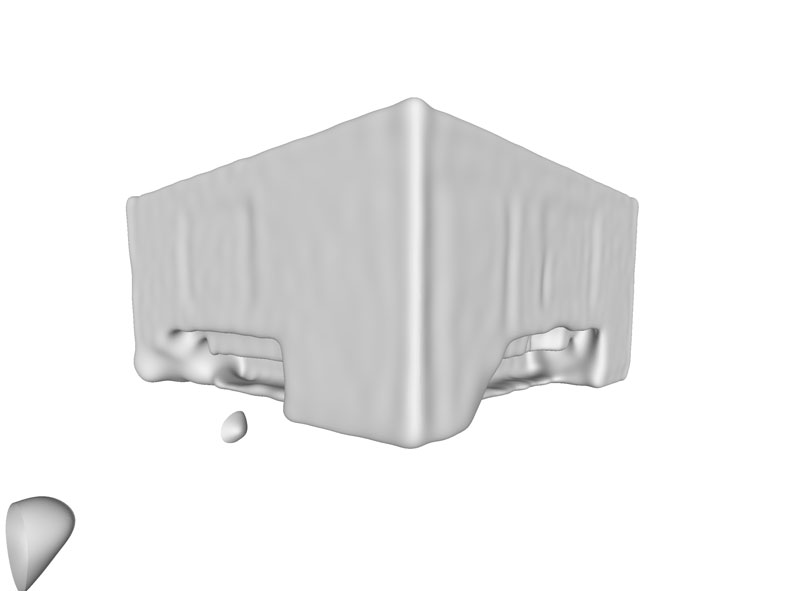}
		\end{minipage}
	}
	\subfloat[Re-initialization ($10000$ epochs)]{
		\begin{minipage}{0.23\linewidth}
			\centering
			\includegraphics[width=0.95\linewidth]{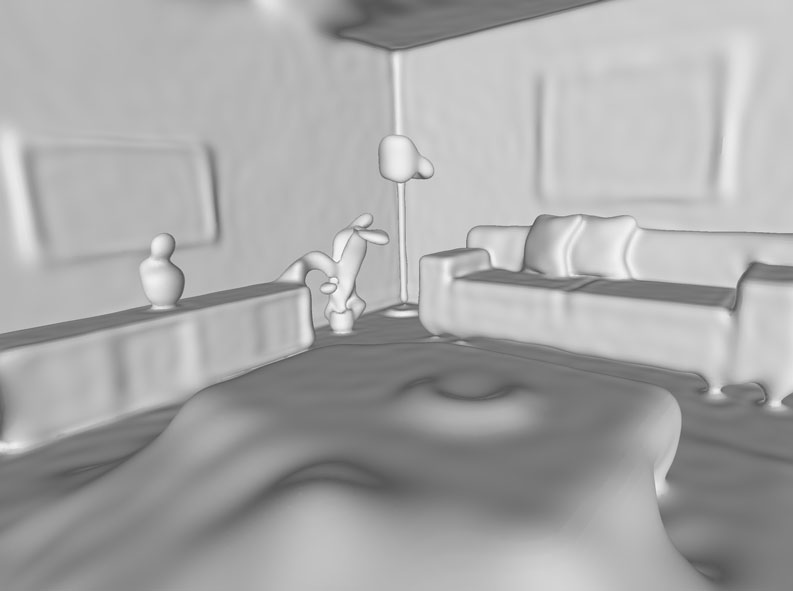}\\
			\includegraphics[width=0.95\linewidth]{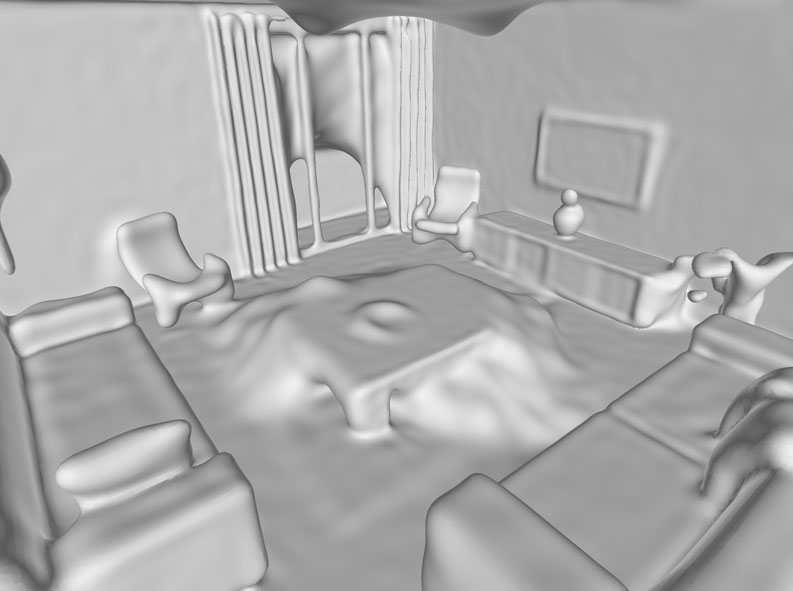}\\
			\includegraphics[width=0.95\linewidth]{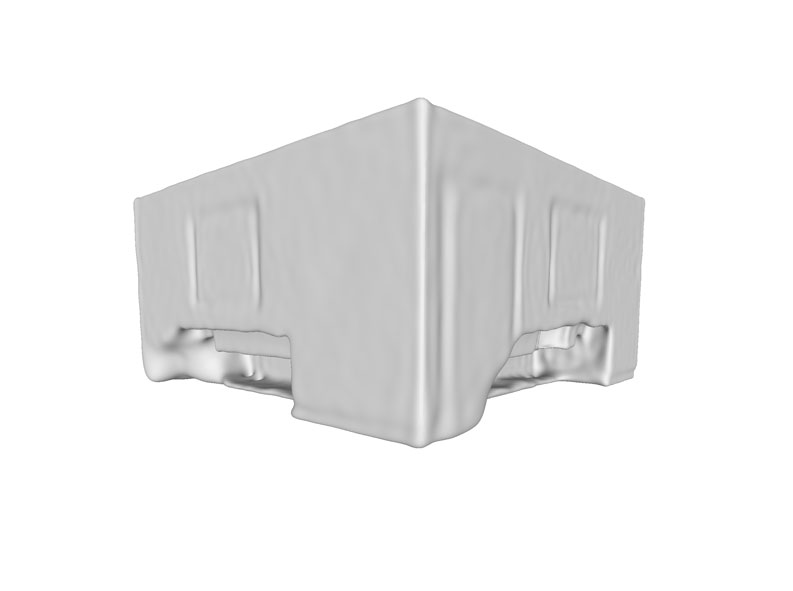}
		\end{minipage}
	}
	\subfloat[Re-training]{
		\begin{minipage}{0.23\linewidth}
			\centering
			\includegraphics[width=0.95\linewidth]{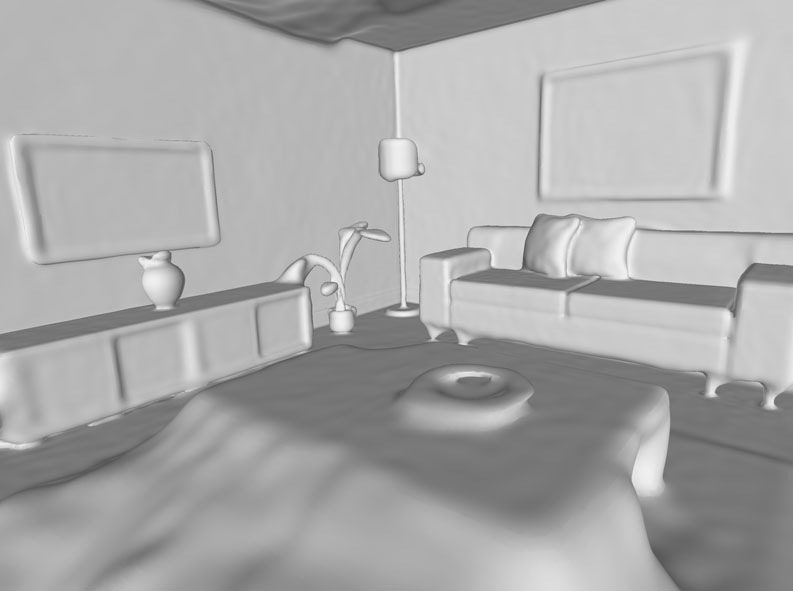}\\
			\includegraphics[width=0.95\linewidth]{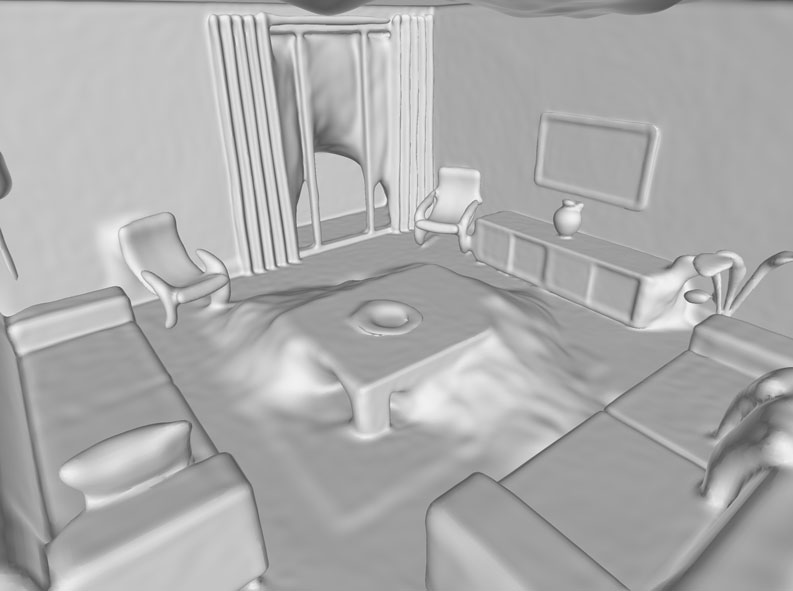}\\
			\includegraphics[width=0.95\linewidth]{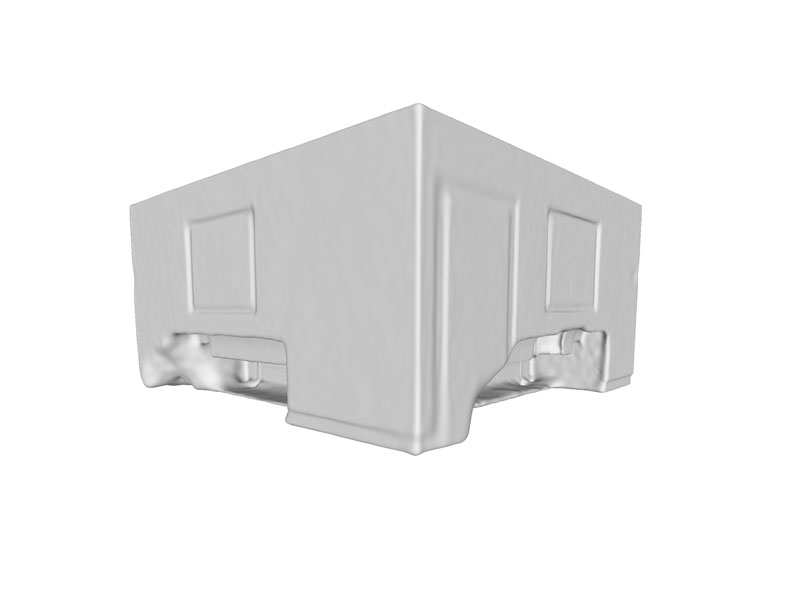}
		\end{minipage}
	}
	\subfloat[Ours]{
		\begin{minipage}{0.23\linewidth}
			\centering
			\includegraphics[width=0.95\linewidth]{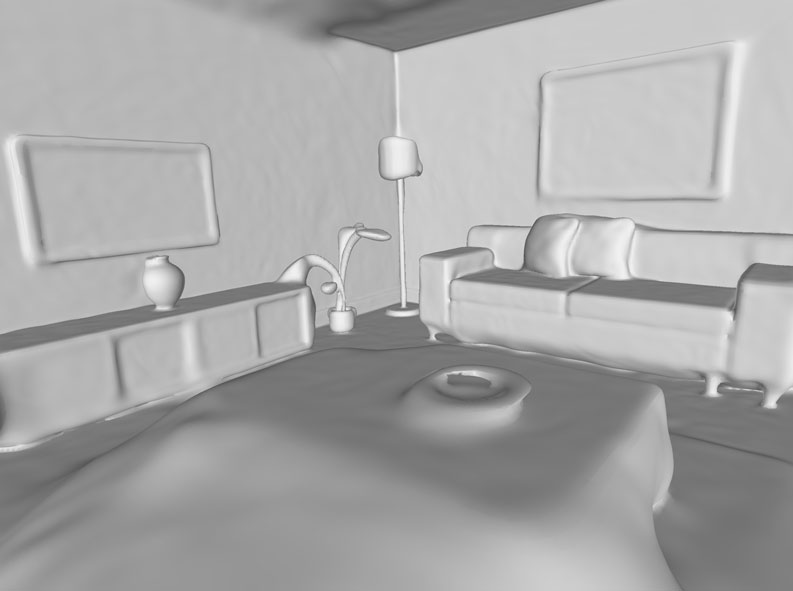}\\
			\includegraphics[width=0.95\linewidth]{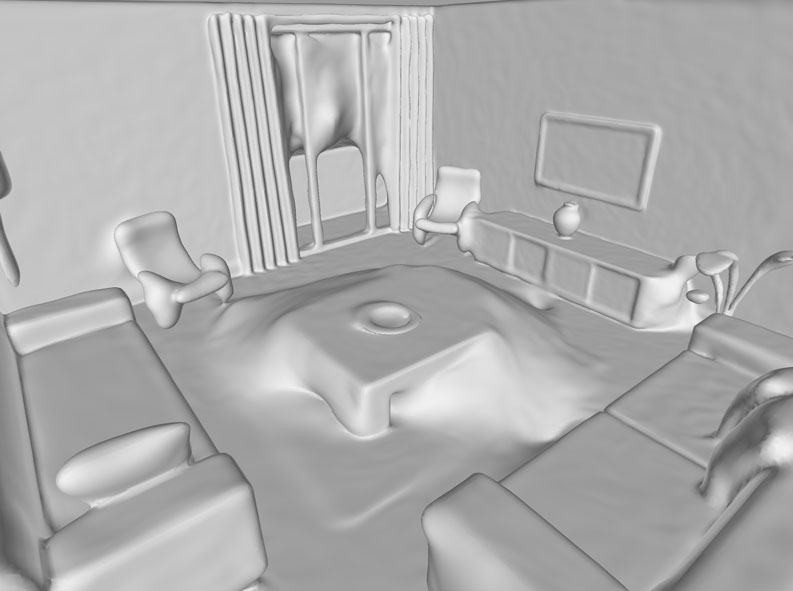}\\
			\includegraphics[width=0.95\linewidth]{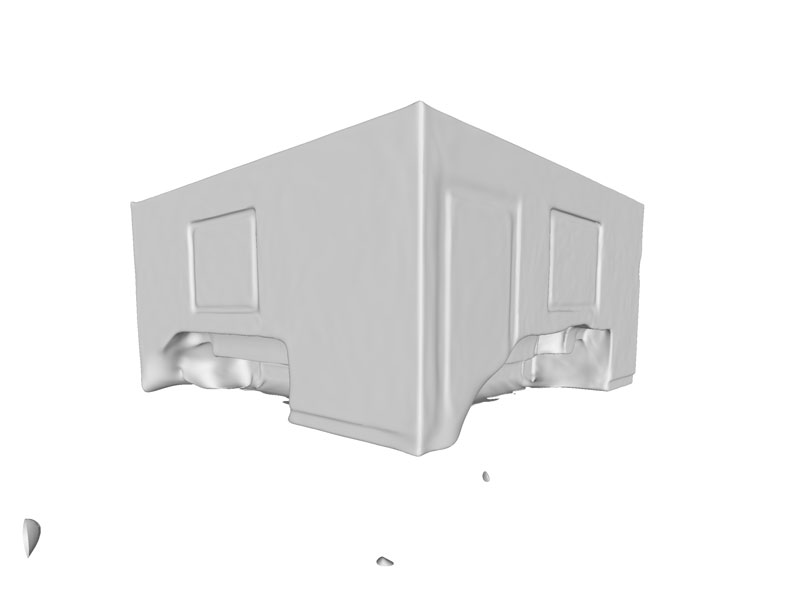}
		\end{minipage}
	}
	\caption{The extracted mesh using approximated SDF values. The proposed approach achieves comparable results against the computationally expensive re-training baseline and outperforms the re-initialization baseline. High-frequency details of the scene geometry cannot be well-recovered by re-initialization baseline even if it is trained for $10000$ epochs.}
	\label{fig:reinit}
\end{figure*}

We first study the effect of network initialization by comparing it against the re-initialization baseline. As illustrated in Fig.~\ref{fig:reinit_loss}, the knowledge distillation through weight initialization leads to faster convergence and better results when a new frame arrives. Fig.~\ref{fig:reinit} displays that the performance of re-initialization baseline deteriorates significantly due to the lack of parameter initialization from previous network parameters. It is noteworthy that the re-initialization baseline cannot recover comparable high-frequency details even after $10000$ epochs of training.

\begin{figure}[htbp]
	\centering
	\includegraphics[width=0.45\linewidth]{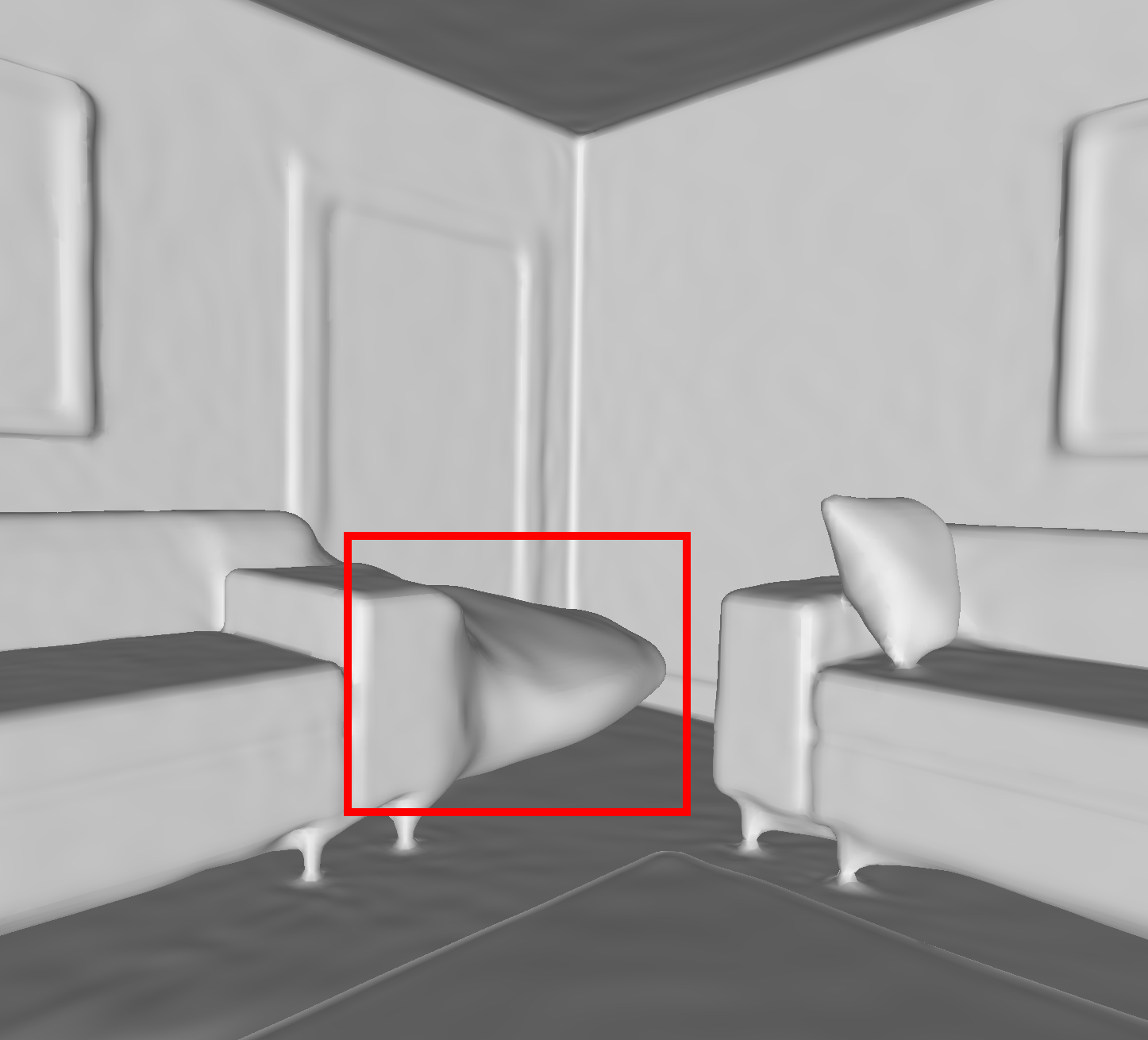}
	\hspace{0.03\linewidth}
	\includegraphics[width=0.45\linewidth]{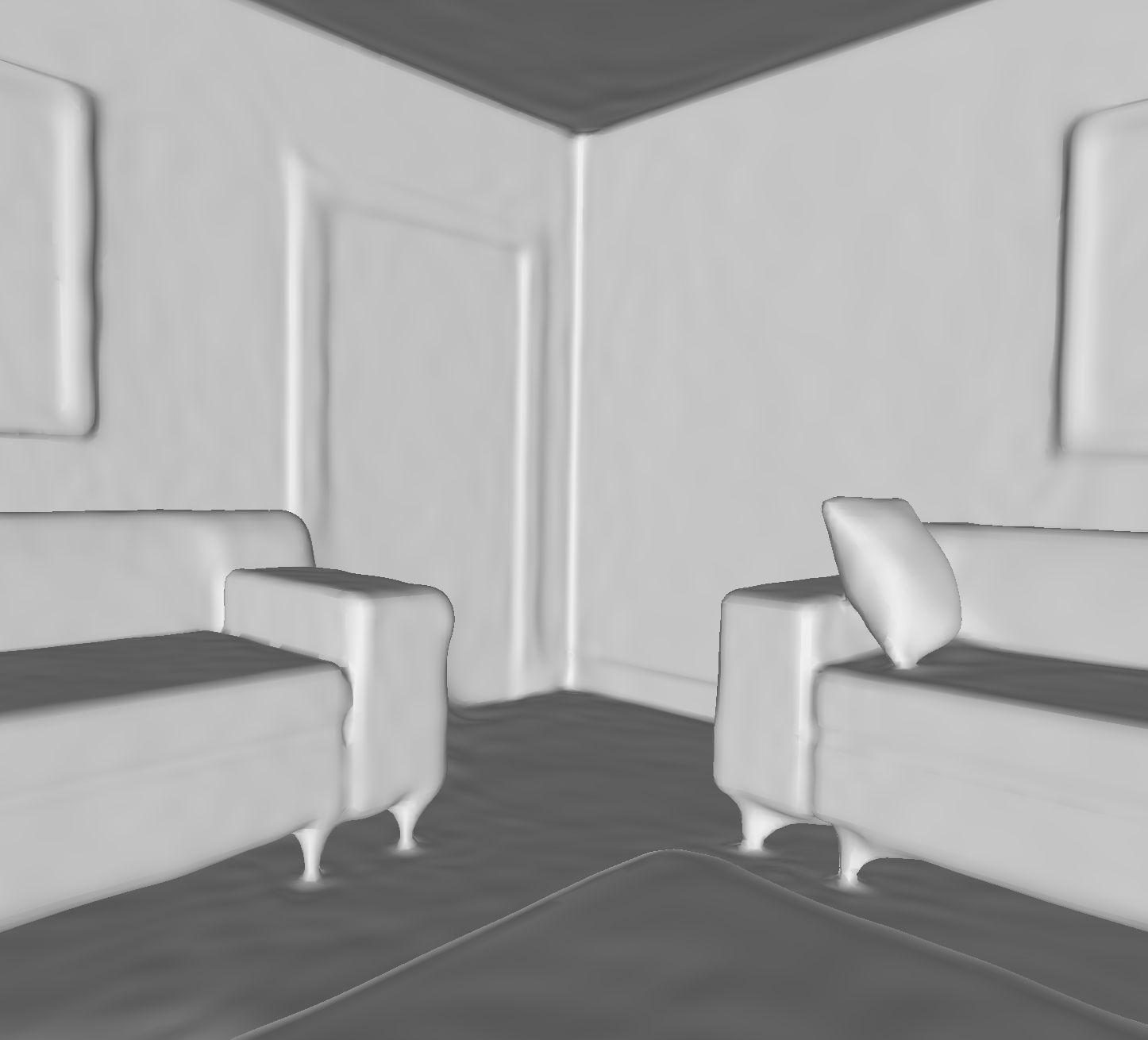}
	\caption{Without the guidance of the last network, false negative SDF  approximation arising from partial occlusion may generate spurious zero level-set surface (left). We exploit past experience to alleviate this issue (right).}
	\label{fig:false_negative}
\end{figure}

On the other hand, the guided sign regularization is crucial to eliminate the false negative distance field arising from occlusion (see Fig.~\ref{fig:sign}). As illustrated in Fig.~\ref{fig:false_negative}, off-surface samples guided by past experience serve as a reliable regularization to constrain the sign of the distance function.

We can also experimentally find that the simple solution of storing a fixed number of buffer with the same size of each frame \cite{Lee2020iclr} is effective enough to serve as a replayed experience of zero level-set observations. Fig.~\ref{fig:heatmap}, \ref{fig:sdf_vis} and the supplementary material demonstrate that training a network without replayed buffer leads to catastrophic forgetting.

\noindent\textbf{Tradeoffs between accuracy and efficiency.} 
Our method guarantees constant training time for each frame (approximately $6$ minutes for $1500$ epochs) due to the fixed size of the replayed buffer. Though an additional memory is required to store the buffer data, the training time will not sacrifice as the batch size are equally divided and attributed to the current data and the buffer data at each iteration. On the contrary, batch re-training baseline will take data from the first frame to the last frame as the entire batch dataset, leading to linearly scaled training time arising from the constantly augmented training data. As illustrated in Fig.~\ref{fig:cover}, when the $\#87$ frame arrives, the batch re-training baseline will take about $13$ hours to train the entire dataset for $1500$ epochs. Hence, we obtain orders of magnitude smaller training time with comparable accuracy when compared with the batch re-training baseline. The storage of past observations is also orders of magnitude smaller. On the other hand, when compared to the fine-tuning and re-initialization baselines, we achieve better accuracy by exploiting the guidance of past experience. The proposed continual neural mapping ensures the incremental network parameter updating in a globally consistent way, achieving a nice trade-off between efficiency and accuracy when compared with alternative baselines. 

\begin{figure}[tbp]	 
	\centering
	\includegraphics[width=0.98\linewidth]{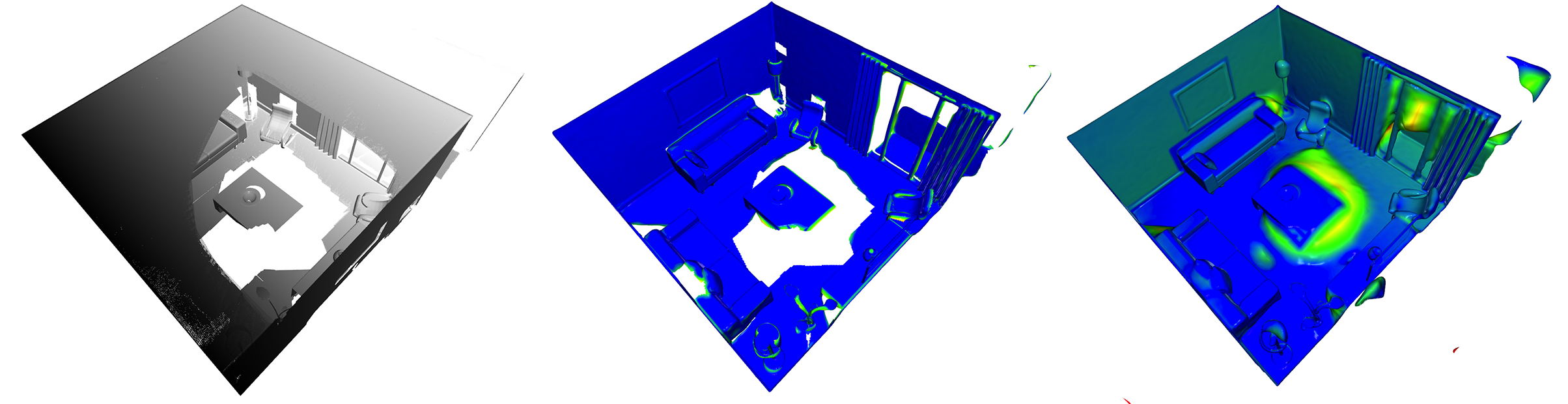}
	\caption{Error map of the extracted mesh model. The majority of erroneous surface lie in the unseen areas (right). By specifying voxel indices for mesh extraction according to the observations as masked model, the accuracy of the mesh outperforms competitives (middle).}
	\label{fig:map_confidence}
\end{figure}

\noindent\textbf{Comparisons against state-of-the-art.} 
\begin{figure*}[htbp]
	\centering
	\includegraphics[width=0.98\linewidth]{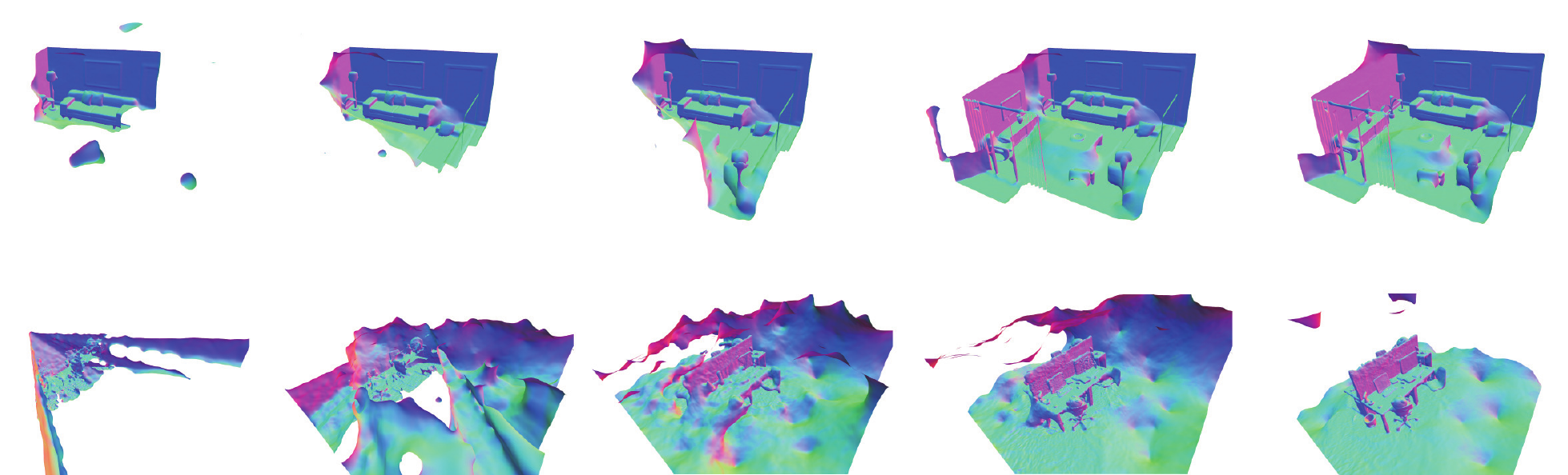}
	\caption{Incrementally updated geometry on the synthetic ICL (top) and real TUM (bottom) datasets. The mesh is visualized with the vertex normal. We refer readers to the supplementary materials for more details.}
	\label{fig:increment}
\end{figure*}
We also compare against the state-of-the-art methods in terms of the maintained model size and the extracted mesh accuracy. For RoutedFusion~\cite{Weder2020cvpr}, we use a voxel size of $2$cm, corresponding to a grid resolution of $512^3$ allocated voxels. For LIG~\cite{Jiang2020cvpr}, we use a part size of $25$cm to meet the point density. As shown in Fig.~\ref{fig:map_confidence}, the continuous characteristic of our signed distance function leads to spurious zero level-set surfaces in unseen areas, hence resulting in low reconstruction accuracy. However, if we follow the competitive methods to reason the occupancy status of each voxel grid according to previous observations and only extract triangle meshes around occupied voxels,  the overall accuracy outperforms the state-of-the-art methods (Tab.~\ref{tab:sota}). It should be noted that we only maintain a single network with a size of less than $800$ KB to achieve $mm$  level of accuracy through continual learning. This is consistent with our average accuracy of SDF approximation in Fig.~\ref{fig:heatmap}.

\begin{table}
	\centering
	\caption{The parameter size of the representations and the cloud/mesh distance error (m) of generated mesh models.}
	\begin{tabular}{lccc}
		\hline
		Method & Mean & Std. & Parameters\\
		\hline
		RoutedFusion~\cite{Weder2020cvpr} & 0.0403& 0.0687 & 512$^3$\\
		LIG~\cite{Jiang2020cvpr} & 0.0106& 0.0146 & 69,795$\times$32\\
		Ours & 0.0584&0.2115 & 198,657\\
		Ours (masked) & \textbf{0.0044}& \textbf{0.0010} & 198,657\\
		\hline
	\end{tabular}
    \label{tab:sota}
\end{table}

The continuous nature of the implicit representation discards the necessity of voxelization, thus guaranteeing a much compact and expressive representation. The contribution of the proposed approach can also be understood from the fusion perspective: Instead of maintaining the discretized value of SDF $\bm{y}$, we resort to the parameter space of a continuous signed distance function. The volumetric fusion of SDF value is replaced by the incremental updating of the network parameters that are learned continually from sequential observations. As illustrated in Fig.~\ref{fig:increment}, accurate and smooth surfaces can be extracted from the incrementally updated network on both synthetic and real datasets.


\section{Conclusion}
In this paper, we introduce a novel \emph{continual neural mapping} problem, aiming to bridge the gap between the prevalent batch-trained implicit neural representation and the commonly used streaming data for robotics and vision applications. We primarily aim to discern if a continual learning solution can eliminate the need for batch data preservation and re-training fashion without catastrophic forgetting for coordinate-based MLP. The answer is positive. Dealing with the SDF regression problem, continual neural mapping benefits from the guidance of past experience and enables a single network to model the scene geometry incrementally from sequential observations. This brings great potentials to tasks with online requirements. Besides, the general problem setting turns scene understanding into an incremental map-centric fashion. 

Exploiting the expressiveness of the neural network as a memory for past sequential observations or a predictor for future exploration may be the route to exciting future work. Potential directions based on the continual neural mapping paradigm include how to achieve faster convergence for real-time applications, how to encode multiple scene properties within a single network continually, and how to enhance the expressiveness and the prediction quality with different network architectures and learning techniques.

\paragraph{Acknowledgements}
We thank anonymous reviewers for their fruitful comments and suggestions. This work is supported by the National Key Research and Development Program of China
(2017YFB1002601) and National Natural Science Foundation of China (61632003, 61771026).
\newpage
{\small

\bibliographystyle{ieee_fullname}
}

\newpage

\dictchar{Supplementary Materials}

\beginsupplement
\section{Introduction}
We refer readers to the video material for better visualizing the dynamical changes of the network, the corresponding SDF approximation, and the extracted mesh models. Here, we provide additional experimental results to complement the experiment section of our main paper. We also outline revenues for interesting future work (highlighted in bold font) through the supplementary experiments.

\section{Baseline Settings}
\begin{figure}[htbp]
	\centering
	\includegraphics[width=0.98\linewidth]{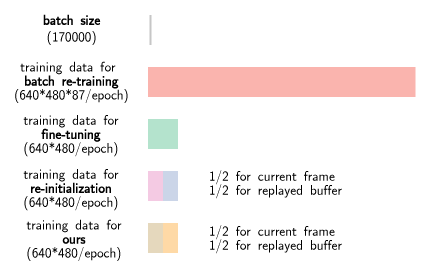}
	\vspace{1em}
	\caption{Training data size for different baselines when the $\#870$ frame arrives. Note that we downsample the data every ten frames. Hence, there are $86$ frames that have been seen before the $\#870$ frame.}
	\vspace{1em}
	\label{fig:data_size}
\end{figure}

\begin{figure}[htbp]
	\centering
	\subfloat[Batch training (mean: 2.06mm, std.: 3.38mm)]{
		\begin{minipage}[t]{0.98\linewidth}
			\centering
			\includegraphics[width=0.98\linewidth]{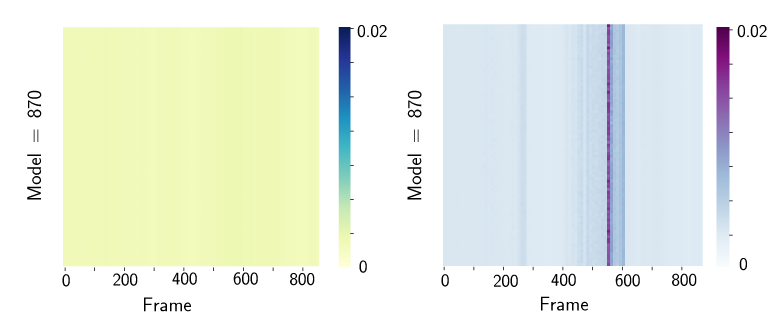}
		\end{minipage}
	}\\ 
	\subfloat[Fine-tuning (mean: 496.34mm, std.: 626.92mm)]{
		\begin{minipage}[t]{0.98\linewidth}
			\centering
			\includegraphics[width=0.98\linewidth]{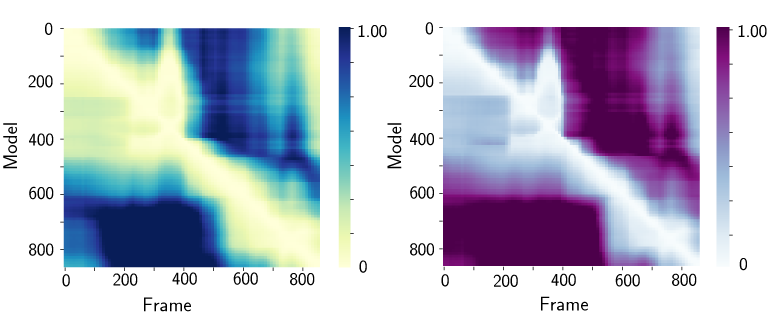}
		\end{minipage}
	}\\ 
	\subfloat[Ours (mean: 2.58mm, std.: 4.22mm)]{
		\begin{minipage}[t]{0.98\linewidth}
			\centering
			\includegraphics[width=0.98\linewidth]{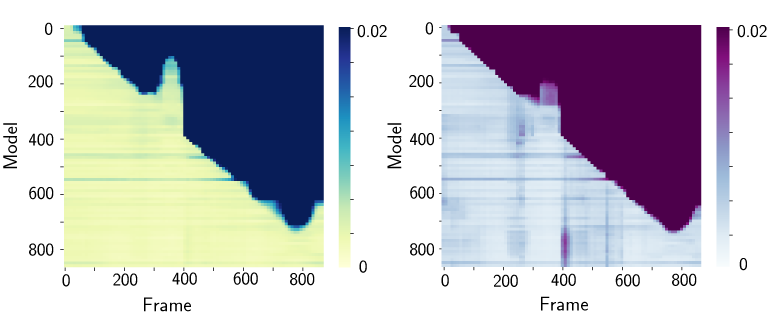}
		\end{minipage}
	} 
	\caption{The heatmap indexed by $(m,n)$ presents the mean (left) and standard deviation (right) of the approximated SDF accuracy $f(\bm{x}^m;\theta^n)$. Notice that the range of fine-tuning baseline differs from others for better visualization. The mean and std. are calculated with values in the lower triangle to measure the accuracy of memory.}
	\label{fig:heatmap_s}
\end{figure}

A more illustrative explanation of the proposed baseline setting is presented in this section. As presented in Fig.~\ref{fig:data_size}, the data size for batch re-training is $87$ times as large as that of ours, thus leading to much more iterations within each epoch (and more training time accordingly). As is presented in Sec.~\ref{subsec:analysis} of the main paper, we achieve nice trade-offs between accuracy and efficiency with much less training time and data storage over the batch re-training baseline (comparable accuracy) and much better accuracy compared to other alternatives (same training time).

It can also be understood from Fig.~\ref{fig:data_size} that the training time of batch re-training will be linearly growing with the increasing number of frames. Ours, on the other hand, maintain a constant training time and memory consumption without forgetting.

\section{Memory and Predictor}
\label{sec:heatmap}
\begin{figure*}[htbp]
	\centering
	\includegraphics[width=0.98\linewidth]{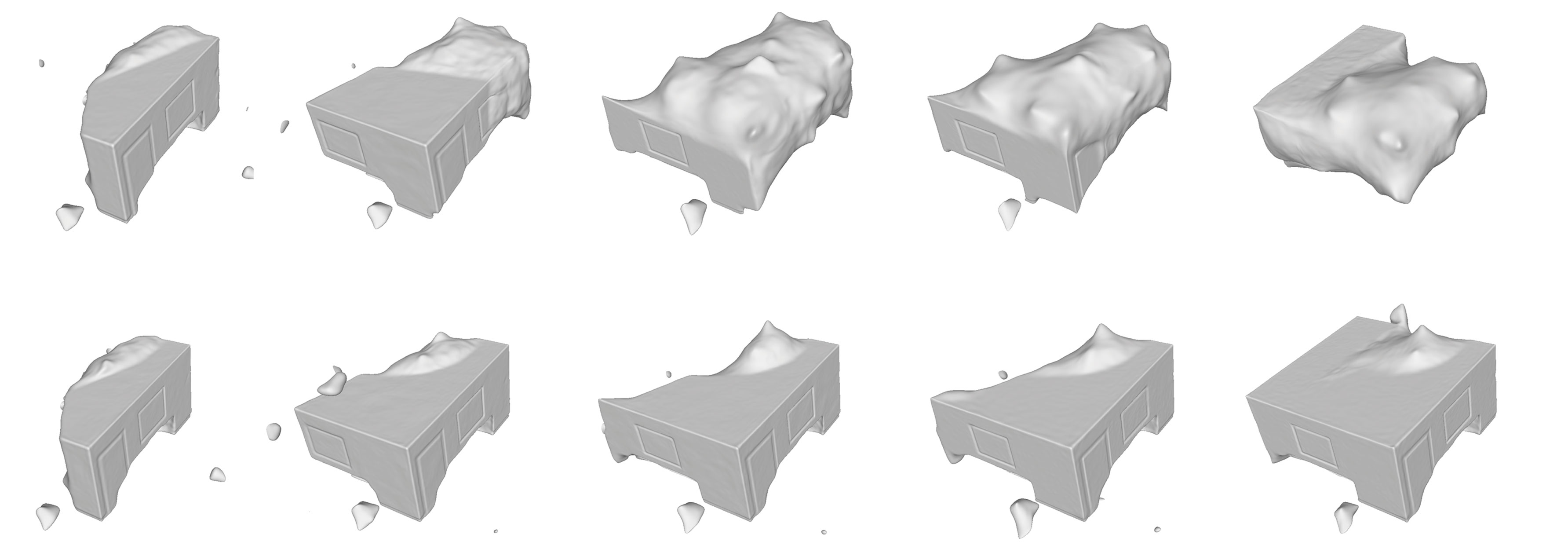}
	\caption{Recovered scene geometry from the implicit mapping function at frame 100-500. Top row: without the replayed buffer to regularize previously visited areas, the network tends to forget the geometry gradually; Bottom row: a simple solution of maintaining a fixed size of buffer can effectively preserve pre-visited scene geometry with high-frequency details.}
	\vspace{1em}
	\label{fig:no_buffer}
\end{figure*}
We here provide a more thorough analysis of Fig.~\ref{fig:heatmap} of our main paper. As illustrated in Fig.~\ref{fig:heatmap_s}, the heat map indexed by $(m,n)$ presents the mean and the standard deviation in meter of the SDF $f(\bm{x}^m_i;\theta^n)$ approximated using the network $\theta^n$ for the observation $\bm{x}^m$ (batch training baseline is trained with the entire sequence of data). \emph{The upper triangle denotes the prediction performance for unseen areas as ${m> n}$ (not applicable for batch training baseline), while the lower triangle denotes the memory performance for previously seen areas as ${m< n}$.} We can see that the proposed method achieves comparable results of memory against the batch re-training baseline, while the fine-tuning baseline suffers from catastrophic forgetting. 

The heatmap also demonstrates the forward transfer (→) and backward transfer (↓) performance~\cite{Delange2021pami} of each method. It is clear that at this stage, a simple MLP does not perform well for predicting unseen areas. \textbf{The incorporation of geometry prior for better prediction} may be an interesting follow-up suggestion.

\section{Analysis of the Replayer Buffer}
We provide additional 3D visualization of the scene geometry changes over time as a complement to Fig.~\ref{fig:sdf_vis} of our main paper, depicting the role of the replayed buffer. As illustrated in Fig.~\ref{fig:no_buffer}. the replayed buffer of zero level-set samples properly regularizes the previously visited surface information and alleviates the catastrophic forgetting issue. Further experiments can be conducted to \textbf{analyze the relationship between the convergence rate and the sample selection strategy}. It would be the key to achieving real-time performance based on the continual neural mapping paradigm.

\section{Experiments on the TUM Dataset}

\begin{figure*}[t]
	\centering
	\begin{minipage}[t]{0.19\linewidth}
		\centering
		\includegraphics[width=0.98\linewidth]{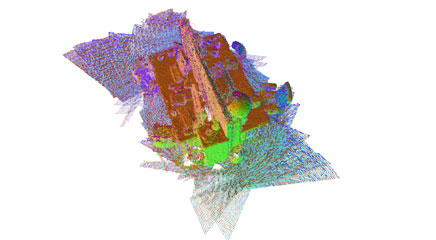}
	\end{minipage}
	\begin{minipage}[t]{0.19\linewidth}
		\centering
		\includegraphics[width=0.98\linewidth]{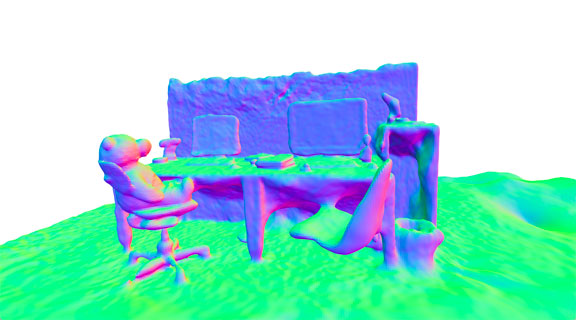}
	\end{minipage}
	\begin{minipage}[t]{0.19\linewidth}
		\centering
		\includegraphics[width=0.98\linewidth]{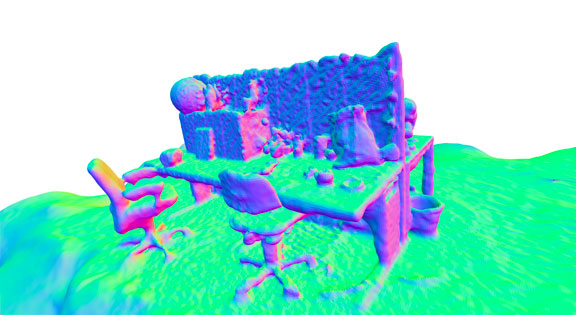}
	\end{minipage}
	\begin{minipage}[t]{0.19\linewidth}
		\centering
		\includegraphics[width=0.98\linewidth]{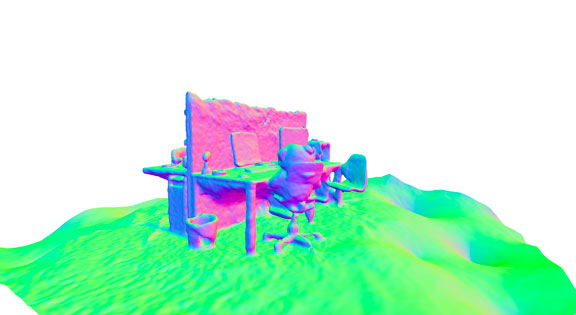}
	\end{minipage}
	\begin{minipage}[t]{0.19\linewidth}
		\centering
		\includegraphics[width=0.98\linewidth]{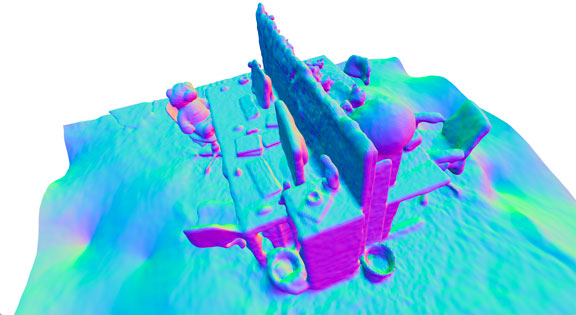}
	\end{minipage}
	\vspace{1em}
	\caption{Extracted mesh (right) with the network that continually learned till the last frame. The mesh is visualized with the vertex normal to present the smooth surface even if the network is learned from noisy sequential data (left).}
	\label{fig:TUM_1}
\end{figure*}

\begin{figure*}[t!]
	\centering
	\includegraphics[width=0.98\linewidth]{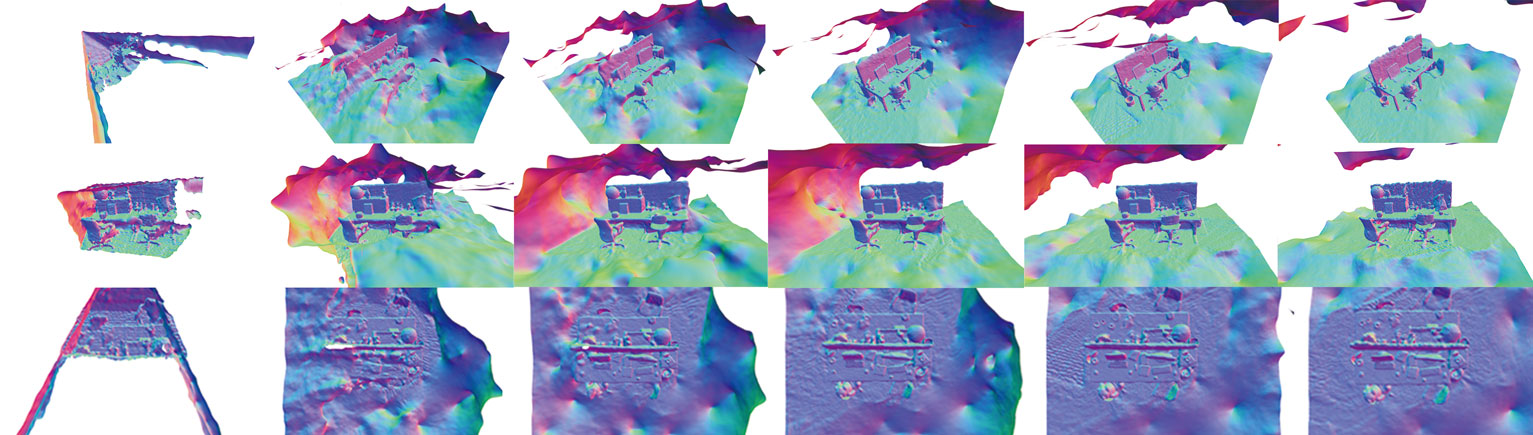}
	\caption{Self-improved SDF approximation on TUM \emph{long office} dataset. Each row is recorded at a specific view point. Each column denotes the mesh extracted from the network trained with the frame $t$ ($t = 30, 480, 930, 1380, 1830$ respectively). We refer readers to the supplementary video for better visualization.}
	\label{fig:TUM_2}
\end{figure*}

\begin{figure*}[h!]
	\centering
	\subfloat[RoutedFusion]{
		\begin{minipage}[t]{0.24\linewidth}
			\centering
			\includegraphics[width=0.95\linewidth]{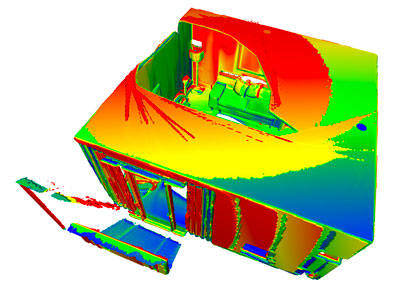}
		\end{minipage}
	}
	\subfloat[LIG]{
		\begin{minipage}[t]{0.24\linewidth}
			\centering
			\includegraphics[width=0.95\linewidth]{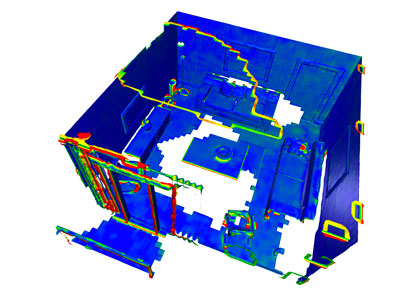}
		\end{minipage}
	}
	\subfloat[Ours]{
		\begin{minipage}[t]{0.24\linewidth}
			\centering
			\includegraphics[width=0.95\linewidth]{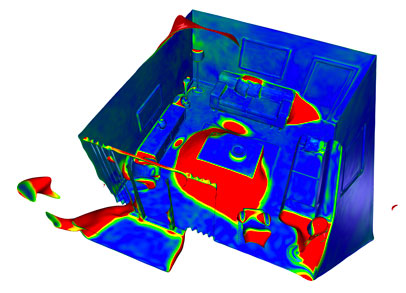}
		\end{minipage}
	}
	\subfloat[Ours (masked)]{
		\begin{minipage}[t]{0.24\linewidth}
			\centering
			\includegraphics[width=0.95\linewidth]{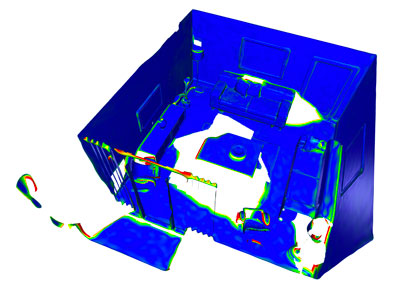}
		\end{minipage}
	} 
	\caption{The visualization of mesh accuracy for each method (scaled up to 5cm, red: low accuracy, blue: high accuracy). We here provide the back face visualization to see the internal error distribution. RoutedFusion generates thick faces. Therefore, the internal view of the room cannot be seen. It is clear that our \emph{continuous SDF approximation} leads to low accuracy in \emph{previously unseen} areas, while the geometry recovered in already seen areas (masked) outperforms the state-of-the-art methods.}
	\label{fig:error_map}
\end{figure*}
We provide additional qualitative results on the real-world TUM dataset~\cite{TUM} from different viewpoints. We refer readers to the supplementary video for better visualizing the recovered model. As illustrated in Fig.~\ref{fig:TUM_1} and~\ref{fig:TUM_2}, we can see that the proposed continual neural mapping setting can mitigate the side effect of noisy observations. However, it is also noteworthy that high-frequency signal recovery and denoising are controversial. \textbf{How to use a single network to recover fine-grained geometry details with sensor noise reduced continually} is one interesting extended case for continual neural mapping.

\section{Error Map of Extracted Mesh Model}
As illustrated in Fig.~\ref{fig:error_map}, we here present the visualization of the extracted mesh accuracy\footnote{CloudCompare: \url{http://www.danielgm.net/cc/}} of~\cite{Weder2020cvpr, Jiang2020cvpr}. The parameter settings for the two methods are specified as follows: for RoutedFusion~\cite{Weder2020cvpr}\footnote{RoutedFusion. https://github.com/weders/RoutedFusion}, we use a voxel size of $2$cm with $512^3$ voxels to best fit the entire scene within the volumetric field using a NVIDIA GeForce RTX 2080Ti; for LIG~\cite{Jiang2020cvpr}\footnote{LIG. https://github.com/tensorflow/graphics}, we set a part size of $0.25$ as suggested by the paper. Though the proposed implicit representation does not rely on the discretized volume, we can maintain active volume indices to extract clean mesh triangles with state-of-the-art accuracy. \textbf{The incorporation of priors to eliminate spurious zero level-set due to incomplete observations} is worth studying (similar to Sec.~\ref{sec:heatmap}).
\end{document}